\documentclass[letterpaper]{article} % DO NOT CHANGE THIS
\usepackage[]{aaai25}  % DO NOT CHANGE THIS
\usepackage{times}  % DO NOT CHANGE THIS
\usepackage{helvet}  % DO NOT CHANGE THIS
\usepackage{courier}  % DO NOT CHANGE THIS
\usepackage[hyphens]{url}  % DO NOT CHANGE THIS
\usepackage{graphicx} % DO NOT CHANGE THIS
\usepackage{natbib}  % DO NOT CHANGE THIS AND DO NOT ADD ANY OPTIONS TO IT
\usepackage{caption} % DO NOT CHANGE THIS AND DO NOT ADD ANY OPTIONS TO I
\usepackage{algorithm}
\usepackage{algorithmic}

\usepackage{amssymb}
\usepackage{amsmath}
\usepackage{mathtools}

\usepackage{newfloat}
\usepackage{listings}
\DeclareCaptionStyle{ruled}{labelfont=normalfont,labelsep=colon,strut=off} % DO NOT CHANGE THIS
\floatstyle{ruled}
\newfloat{listing}{tb}{lst}{}
\floatname{listing}{Listing}

\usepackage{booktabs}
\usepackage{makecell}
\usepackage{multirow}

\usepackage{tabularx}
\newcolumntype{Y}{>{\centering\arraybackslash}X} % Define a new type Y which makes the content in the table be center aligned

\usepackage{enumitem}
\usepackage{cleveref}
\usepackage{xspace}
\usepackage{longtable}

\crefname{table}{Table}{Tables}
\crefname{figure}{Figure}{Figures}
\crefname{section}{Section}{Section}

\def\eg{\emph{e.g.}\xspace} 
\def\ie{\emph{i.e.}\xspace} 
\def\etc{\emph{etc.}\xspace}

\newcommand{\ours}{MIFAG\xspace}
\newcommand{\dataset}{MIPA\xspace}

\title{Learning 2D Invariant Affordance Knowledge for 3D Affordance Grounding}
\author {
    Xianqiang Gao\equalcontrib\textsuperscript{\rm 1,\rm 2},
    Pingrui Zhang\equalcontrib\textsuperscript{\rm 2,\rm 3},
    Delin Qu\textsuperscript{\rm 2,\rm 3}\\
    Dong Wang\textsuperscript{\rm 2},
    Zhigang Wang\textsuperscript{\rm 2},
    Yan Ding\textsuperscript{\rm 2},
    Bin Zhao\textsuperscript{\rm 2}\thanks{Corresponding author}
}
\affiliations {
    \textsuperscript{\rm 1}University of Science and Technology of China\\
    \textsuperscript{\rm 2}Shanghai AI Laboratory\\
    \textsuperscript{\rm 3}Fudan University\\
    gaoxianqiang@mail.ustc.edu.cn, \{zhangpingrui,zhaobin\}@pjlab.org.cn
}

\begin{document}

\maketitle

\begin{abstract}

3D Object Affordance Grounding aims to predict the functional regions on a 3D object and has laid the foundation for a wide range of applications in robotics. Recent advances tackle this problem via learning a mapping between 3D regions and a single human-object interaction image. However, the geometric structure of the 3D object and the object in the human-object interaction image are not always consistent, leading to poor generalization. To address this issue, we propose to learn generalizable invariant affordance knowledge from multiple human-object interaction images within the same affordance category. Specifically, we introduce the \textbf{M}ulti-\textbf{I}mage Guided Invariant-\textbf{F}eature-Aware 3D \textbf{A}ffordance \textbf{G}rounding (\textbf{MIFAG}) framework. It grounds 3D object affordance regions by identifying common interaction patterns across multiple human-object interaction images. First, the Invariant Affordance Knowledge Extraction Module (\textbf{IAM}) utilizes an iterative updating strategy to gradually extract aligned affordance knowledge from multiple images and integrate it into an affordance dictionary. Then, the Affordance Dictionary Adaptive Fusion Module (\textbf{ADM}) learns comprehensive point cloud representations that consider all affordance candidates in multiple images. Besides, the Multi-Image and Point Affordance (\textbf{MIPA}) benchmark is constructed and our method outperforms existing state-of-the-art methods on various experimental comparisons. Project page: \url{https://goxq.github.io/mifag}

\end{abstract}

\section{Introduction}

3D Object Affordance Grounding seeks to identify and predict functional regions on an object's 3D point cloud. This task has laid the foundation for connecting visual perception with physical operation and paved the way for a wide range of real-world applications, such as embodied systems~\cite{ahn2022i,driess2023palm,wu2023tidybot}, object manipulation~\cite{huang2024a3vlm,huang2023instruct2act,wu2021vat,li2024simultaneous} and object grasping~\cite{dai2023graspnerf}.

Currently, methods in 3D object affordance prediction can be divided into two categories. One of them involves utilizing reinforcement learning, which requires an agent to interact with objects repeatedly in a simulated environment. Such methods are usually time-consuming, and their generalization performance is often restricted by the limited number of simulation scenes~\cite{mo2021where2act,ning2024where2explore,cheng2023learning}. The other category leverages labeled 3D object affordance data and attempts to learn a mapping between objects and affordance in an end-to-end manner. This kind of approach is more direct and has gained more and more popularity recently. Specifically, 3D AffordanceNet~\cite{deng20213d} first constructs a 3D point cloud affordance grounding benchmark. After that, to accommodate the diversity of object affordances and incorporate human interaction references, many methods have been proposed, either combining a single image or textual description with the point cloud for 3D affordance prediction~\cite{yang2023grounding,li2024laso}.

\begin{figure}[t]
    \centering
    \includegraphics[width=\columnwidth]{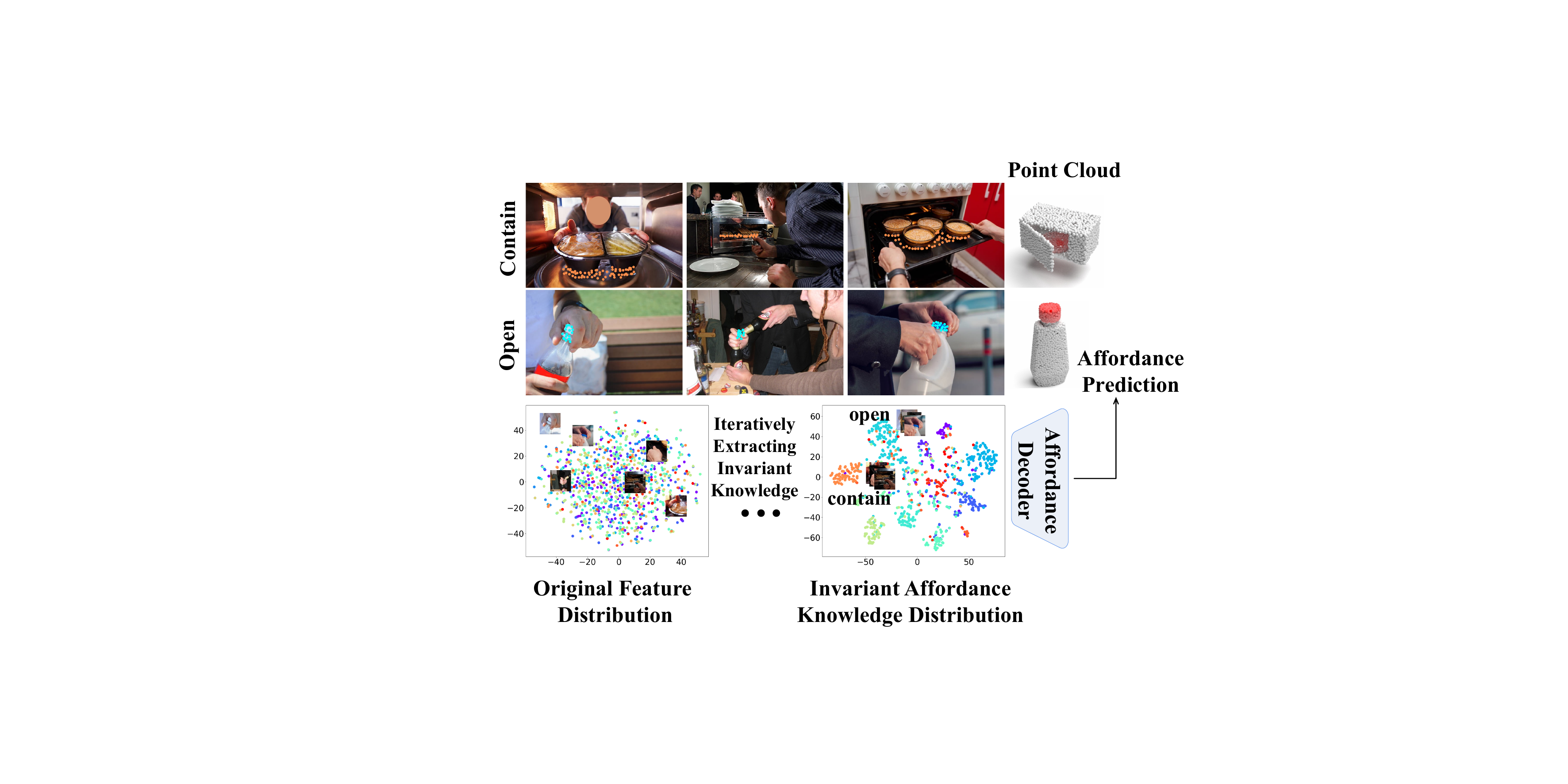}
    \caption{\textbf{Motivation of Our Method.} The reference human-object images exhibit significant variations in appearance, yet they consistently imply the same affordance knowledge. We propose to iteratively extract the invariant affordance knowledge from multiple images, leading to improved performance.}
    \label{fig:motivation}
\end{figure}

Though pioneer research studies have achieved promising progress, they are still limited by failing to leverage the strong correlations and implied invariant affordance knowledge among objects within the same affordance category. As shown in~\cref{fig:motivation}, the affordance of an object is usually determined by multiple human-object reference images. These real-world images exhibit significant variations in appearance yet represent the same object and affordance category. Thus, they share invariant affordance knowledge and strong internal relationships, providing valuable insights into the object's affordance across different contexts. Similarly, predicting the affordance of an object requires the invariant knowledge derived from numerous human-object interaction images. However, previous approaches have not fully addressed these limitations, often either lacking visual information or simply focusing on mapping 3D regions to a single human-object interaction image, leading to poor generalization and suboptimal performance.

Overcoming the aforementioned limitations is non-trivial due to the following challenges: (1) In real-world scenarios, objects of the same category often exhibit significant variations in their interactive regions. Relying solely on textual descriptions, such as ``open the oven'', is insufficient for conveying affordance knowledge. For instance, ovens from different brands can vary in size and handle position. Additionally, simply combining multiple images as references may not yield the expected results due to the considerable diversity in their appearances (\eg, shape, scale, \etc). Therefore, the first challenge lies in how to effectively utilize multiple images with diverse appearances for affordance guidance and how to extract the invariant affordance knowledge. (2) There is a significant gap between the modalities of reference images and point clouds, presenting another challenge: how to effectively integrate invariant affordance knowledge into the point cloud for accurate affordance prediction.

To address the above challenges, we propose the \textbf{M}ulti-\textbf{I}mage Guided Invariant-\textbf{F}eature-Aware 3D \textbf{A}ffordance \textbf{G}rounding (\textbf{\ours}) framework, which gradually extracts affordance knowledge from multiple human-object reference images and effectively integrates this invariant knowledge with point cloud representations to achieve accurate affordance prediction. Specifically, our proposed \ours consists of two modules: the Invariant Affordance Knowledge Extraction Module (\textbf{IAM}) and the Affordance Dictionary Adaptive Fusion Module (\textbf{ADM}). The IAM progressively extracts invariant affordance knowledge across different images using a multi-layer network, while its dual-branch structure minimizes interference caused by appearance variations in the images. The output of the last layer of IAM forms an affordance dictionary that encapsulates all invariant affordance knowledge from these images. Following this, we design the ADM to fuse the extracted invariant affordance knowledge with point clouds, and use the point cloud to query each candidate in the affordance dictionary, thereby obtaining comprehensive point cloud feature representations that consider all affordance candidates.

Our main contributions can be summarized as follows:

\begin{itemize}
    \item We introduce a novel \ours framework for 3D object affordance grounding, which extracts invariant affordance knowledge from multiple reference images.

    \item We propose the IAM, which progressively extracts invariant affordance knowledge from images while minimizing interference caused by appearance variations. The ADM is then proposed to leverage this knowledge to obtain comprehensive point cloud features that consider all affordance candidates.

    \item We construct the Multi-Image and Point Affordance (\dataset) benchmark to advance research in understanding affordances across diverse visual data. Experimental results demonstrate that our \ours outperforms previous state-of-the-art methods.

\end{itemize}

\section{Related Work}

\begin{figure*}[t!]
    \centering
    \includegraphics[width=\textwidth]{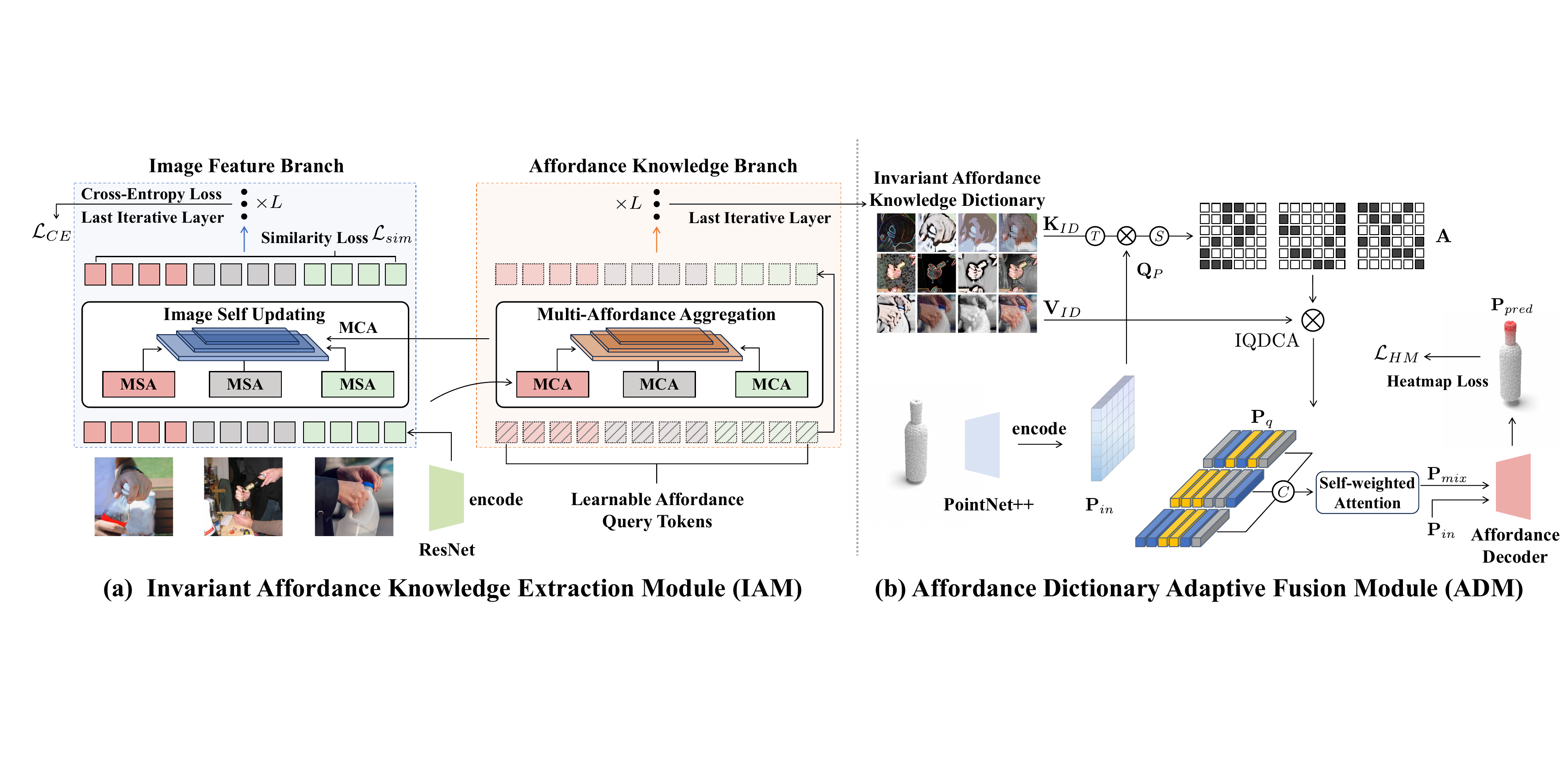}
    \caption{\textbf{Overview of our proposed \ours.} (a) The \textbf{IAM} utilizes a multi-layer network and a dual-branch structure to gradually extract invariant affordance knowledge and minimize interference caused by appearance variations in the images. (b) The \textbf{ADM} leverages the invariant affordance knowledge dictionary derived from (a), using dictionary-based cross attention and self-weighted attention to comprehensively fuse the affordance knowledge with point cloud representations.}
    \label{fig:framework}
\end{figure*}

\subsection{Affordance Learning}
Affordance learning is crucial in robotics, particularly for manipulating articulated objects~\cite{ning2024where2explore}. Consequently, many studies focus on detecting the affordance areas of target objects or scenes~\cite{fang2018demo2vec,luo2022grounded,luo2022learning,nagarajan2019grounded}. Some works extract affordance knowledge from 2D data, \ie, images and videos~\cite{chuang2018learning,do2018affordancenet,Li2023G2L,luo2021one,roy2016multi,thermos2020deep,zhao2020object,chen2023affordance,mur2023multi}, while others explore affordance using natural language guidance~\cite{lu2022phrase,mi2020intention,mi2019object,chen2024worldafford,yoshida2024text}. These methods aim to predict the affordance regions of 2D targets. However, robotics tasks often require 3D information, and transferring affordance knowledge learned from 2D data to 3D scenarios can lead to failures in real-world applications. As a result, several 3D affordance-related datasets have been proposed~\cite{Geng_2023_CVPR,Liu_2022_CVPR,Mo_2019_CVPR}, and some studies focus on leveraging these 3D datasets to ground object affordance~\cite{mo2022o2o,nagarajan2020learning,xu2022partafford,delitzas2024scenefun3d}. 3D AffordanceNet~\cite{deng20213d} first introduces a benchmark dataset for grounding affordance regions on object point clouds. Building on this 3D dataset, IAGNet~\cite{yang2023grounding} proposes a method for learning 3D affordance from reference images. However, the significant appearance variations among reference images and 3D point clouds pose a challenge. LASO~\cite{li2024laso} deals with this by replacing images with natural language descriptions of the interaction area. However, completely discarding the vision information may overlook distinct affordance area guidance. Therefore, we propose a method that better aligns visual latent features from multiple images to effectively guide affordance grounding. Where2Explore~\cite{ning2024where2explore} acquires affordance information by explicitly estimating the geometric similarity across different categories, thereby enhancing generalization. GAPartNet~\cite{Geng_2023_CVPR} introduces cross-category tasks and a dataset to explore the consistency of generalizable and actionable parts. PartManip~\cite{geng2023partmanip} identifies actionable parts to facilitate cross-category object manipulation. While these approaches do not fully capture the category-level consistency features crucial for affordance grounding. By disregarding inconsistent parts across images and implicitly emphasizing consistency, our method more effectively learns affordance information from reference images.

\subsection{Image-Point Cloud Cross-Modal Learning}

Cross-modal tasks enhance individual modalities by incorporating information from one or more other modalities. These additional modalities can serve as conditions to guide the learning progress, often resulting in a positive impact~\cite{qu2024livescene}. The combination of images and point clouds is particularly complementary in downstream tasks: point clouds provide stereospatial information, while images offer rich color and texture details. To leverage this complementarity, many works focus on aligning each pixel to a corresponding point to enrich the semantic information of raw data~\cite{zhuang2021perception,zhao2021lif,xu2022fusionrcnn,vora2020pointpainting,tan2021mbdf,krispel2020fuseseg,chen2024mos}. Additionally, some methods aim to learn multi-view fusion with point clouds, seeking common knowledge and aligning both image-to-image and image-to-point cloud~\cite{jaritz2019multi,zhao2021similarity,li2022bevformer,yang2023bevformer,wang2024driving,jiang2024far3d,chen2024m,zhang2024radocc,xu2024regulating,jiao2024instance}. However, unlike these approaches, which may not fully capture the relationships between multiple images and often lack sufficient modeling of consistent image features for effective fusion with point clouds, our method first establishes consistency between images and then aligns the invariant image features with the point cloud, leading to a more robust integration of visual and spatial information.

\section{Method}

\subsection{Overview}
\label{Section:3.1}

Given a 3D object $\mathbf{P}=\{\mathbf{P}_{c}, \mathbf{P}_{label}\}$ and its corresponding $n$ reference images $\{I_1, I_2,\dots, I_n\}$, where $\mathbf{P}_{c} \in \mathbb{R}^{N \times 3}$ represents the point cloud coordinates, $\mathbf{P}_{label} \in \mathbb{R}^{N \times 1}$ denotes the affordance annotation for each coordinate, and $I_i \in \mathbb{R}^{3 \times H \times W},i=1,\dots,n$, our goal is to ground the affordance region $\mathbf{P}_{pred} \in \mathbb{R}^{N \times 1}$ on the point cloud using the invariant affordance knowledge derived from the $n$ images.

As illustrated in~\cref{fig:framework}, our \textbf{\ours} framework consists of two modules: the Invariant Affordance Knowledge Extraction Module (\textbf{IAM}) and the Affordance Dictionary Adaptive Fusion Module (\textbf{ADM}), which are used to extract invariant affordance knowledge across multiple images and fuse this knowledge with point clouds, respectively. Specifically, in the IAM, a multi-layer network is employed to gradually extract affordance knowledge from images, while a dual-branch structure is designed to minimize interference caused by appearance variations in the images. The output of the last layer of the Affordance Knowledge Branch forms an Invariant Affordance Knowledge Dictionary that encapsulates the affordance knowledge across all reference images. Then, in the ADM, we apply dictionary-based cross-attention and self-weighted attention to fuse the invariant affordance knowledge with the point clouds, resulting in the output $\mathbf{P}_{mix} \in \mathbb{R}^{n \times N \times C}$. Finally, an affordance decoder is used to produce the final affordance prediction $\mathbf{P}_{pred} \in \mathbb{R}^{N \times 1} $.

\subsection{Invariant Affordance Knowledge Extraction Module}
\label{sec:IAM}

The appearances of the object among the reference images vary significantly, yet they all share a common affordance category and provide valuable insights into how the object can be used. Accordingly, we design the IAM to align multiple images based on their common affordance type and to gradually extract the invariant affordance knowledge using a multi-layer network and dual-branch structure.

As illustrated in~\cref{fig:framework} (a), we first randomly initialize a series of Learnable Affordance Query Tokens $\mathbf{Q}$, which are used to represent the learned invariant affordance knowledge from the reference images. These query tokens are then iteratively updated by the multi-layer network using multi-head cross attention, with the reference image features serving as the key and value, as formulated below:
\begin{equation}
\label{Equ:1}
    \mathbf{Q}^{(l)}_{i} = \operatorname{MCA}(\mathbf{Q}^{(l-1)}_{i}\mathbf{W}_{q},\mathbf{F}^{(l-1)}_{i}\mathbf{W}_{k},\mathbf{F}^{(l-1)}_{i}\mathbf{W}_{v}),
\end{equation}
where $\mathbf{Q}^{(l-1)}_{i} \in \mathbb{R}^{M \times C}$ and $\mathbf{F}^{(l-1)}_{i} \in \mathbb{R}^{D \times H \times W}$ denote the affordance queries and reference image features from layer $(l-1)$, respectively. Then, to leverage the inherent consistency among multiple affordance queries, we apply an $\operatorname{MLP}$ layer to align and aggregate all the queries. The aggregated features are then fed back into the image feature branch by multi-head cross attention. Additionally, to continually extract consistent image features, the image feature branch is also updated iteratively using a multi-head self-attention layer. This process can be expressed as follows:
\begin{align}
     \mathbf{Q}^{(l)}_{f} &= \operatorname{MLP}(\mathbf{Q}^{(l)}_1,\mathbf{Q}^{(l)}_2,\dots,\mathbf{Q}^{(l)}_n),\\
     \mathbf{\bar{F}}^{(l-1)}_{i} &= \operatorname{MSA}(\mathbf{F}^{(l-1)}_{i}\mathbf{W}), \\
     \mathbf{F}^{(l)}_{i} &=\operatorname{MCA}(\mathbf{\bar{F}}^{(l-1)}_{i}\mathbf{W}_{q},\mathbf{Q}^{(l)T}_{f}\mathbf{W}_{k},\mathbf{Q}^{(l)}_{f}\mathbf{W}_{v}),
\end{align}
where $\mathbf{Q}^{(l)}_{f},\mathbf{Q}^{(l)}_1, \dots, \mathbf{Q}^{(l)}_n \in \mathbb{R}^{M\times C}$ and $\bar{\mathbf{F}}^{(l-1)}_{i}, \mathbf{F}^{(l)}_{i} \in \mathbb{R}^{D\times H\times W}$. At every iterative layer, to enforce the constraint that all images share the same affordance category, we calculate the similarity loss among all image features $\mathbf{F}^{(l)}_{1}, \mathbf{F}^{(l)}_{2},\dots,\mathbf{F}^{(l)}_{n}$.

In the IAM, the affordance knowledge branch and image feature branch work in tandem. The learnable affordance query tokens iteratively interact with the image features, adding useful affordance information to the image feature branch. Meanwhile, in the affordance knowledge branch, invariant affordance knowledge is gradually aggregated under the guidance of the image feature branch. This dual-branch structure minimizes interference from varying image features on the affordance knowledge. Finally, the output of the last iterative layer of the affordance knowledge branch forms the Invariant Affordance Knowledge Dictionary that encapsulates the affordance knowledge from reference images.

\subsection{Affordance Dictionary Adaptive Fusion Module}
\label{sec:ADM}

\begin{table*}[t!]
\centering
\begin{tabularx}{\textwidth}{c r YYYYYYYYYY}
\toprule
 & \textbf{Metrics} & \textbf{PMF} & \textbf{MBDF} & \textbf{FRCNN} & \textbf{ILN} & \textbf{PFusion} & \textbf{XMF} & \textbf{IAGNet} & \textbf{LASO}   & \textbf{\ours}  \\
\midrule
\multirow{4}{*}{\rotatebox{90}{\textbf{\texttt{Seen}}}}
        & AUC $\uparrow$      & 80.46    & 79.05     & 80.33  & 80.48   & 80.55 & 80.04  & 82.95      & 83.13       & \textbf{85.10} \\
        & aIOU $\uparrow$     & 10.04    & 12.68     & 10.33  & 10.18  & 10.78  & 9.76  & 17.92      & 17.27       & \textbf{20.50} \\
        & SIM $\uparrow$      & 0.445    & 0.476     & 0.449  & 0.447  & 0.449  & 0.442  & 0.547      & 0.540       & \textbf{0.568} \\
        & MAE $\downarrow$    & 0.125    & 0.114     & 0.124  & 0.125  & 0.129  & 0.129  & 0.094      & 0.097       & \textbf{0.091} \\

\midrule
\multirow{4}{*}{\rotatebox{90}{\textbf{\texttt{Unseen}}}}

        & AUC $\uparrow$      & 69.14        & 65.21     & 68.75  & 69.66  & 68.10  & 68.71   & 69.91      & 63.89    & \textbf{71.13}   \\
        & aIOU $\uparrow$     & 3.99         & 4.65      & 3.18   & 4.79   & 4.46  & 3.96   & 5.12       & 5.18     & \textbf{5.23}   \\
        & SIM $\uparrow$      & 0.302        & 0.305     & 0.299  & 0.304  & 0.302  & 0.301   & 0.310      & 0.299    & \textbf{0.315}   \\
        & MAE $\downarrow$    & 0.152        & 0.142     & 0.213  & 0.164  & 0.142  & 0.172   & 0.144      & 0.140    & \textbf{0.136}   \\
\bottomrule
\end{tabularx}
\caption{\textbf{Affordance Prediction Metrics on \textbf{\dataset} Benchmark.} Comparison of evaluation between the proposed method \textbf{\ours} and baseline methods on \textbf{\dataset}. \textbf{\ours} significantly surpasses existing methods and achieves SOTA performance.}
\label{tab:main_results}
\end{table*}

To leverage the Invariant Affordance Knowledge Dictionary obtained in~\cref{sec:IAM}, we design the ADM to adaptively integrate the affordance knowledge into the point cloud representations. As illustrated in~\cref{fig:framework} (b), we first calculate the point-feature weighted query $\mathbf{P}_{q}$. Next, $\mathbf{P}_{q}$ is fused through a self-weighted attention layer to obtain the point-affordance-mixed feature $\mathbf{P}_{mix}$.

Unlike existing multi-head self-attention, which generates query, key, and value tokens from the input feature itself, our approach utilizes the obtained Invariant Affordance Knowledge as an extra dictionary to seamlessly guide affordance learning on the point cloud during the training phase~\cite{Zhang_2024_CVPR}. In line with this, we propose the adaptive Invariant-aware Query Dictionary Cross-Attention (IQDCA). Initially, we use the point cloud feature $\mathbf{P}_{in}$ to generate the query $\mathbf{Q}_P$ for cross-attention, while the keys $\mathbf{K}_{q_1}, \mathbf{K}_{q_2},\dots, \mathbf{K}_{q_n}$ and values $\mathbf{V}_{q_1}, \mathbf{V}_{q_2},\dots, \mathbf{V}_{q_n}$ are obtained from the Invariant Affordance Knowledge Dictionary as follows:
\begin{align}
     \mathbf{Q}_P &= \mathbf{P}_{in} \mathbf{W}_q,\\
     \mathbf{K}_{ID} &= [\mathbf{Q}_1, \mathbf{Q}_2,\dots,\mathbf{Q}_n] \mathbf{W}_k, \\
     \mathbf{V}_{ID} &= [\mathbf{Q}_1, \mathbf{Q}_2,\dots,\mathbf{Q}_n] \mathbf{W}_v, 
\end{align}
where $\mathbf{Q}_P \in \mathbb{R}^{N \times d}, \mathbf{K}_{ID}, \mathbf{V}_{ID} \in \mathbb{R}^{n \times M\times d}$. We then apply the cross-attention mechanism to calculate the dictionary attention matrix $\mathbf{A}$:
\begin{equation}
    \mathbf{A} = \operatorname{SoftMax}(\operatorname{Sim}_{\cos}(\mathbf{Q}_P, \mathbf{K}_{ID})),
\end{equation}
where $\operatorname{Sim}_{\cos}$ denotes the cosine similarity function. The dictionary attention matrix $\mathbf{A} \in \mathbb{R}^{n \times N\times M}$ represents the similarity between the point cloud and the invariant query for each individual image. This matrix serves as guidance to adaptively refine the knowledge in the dictionary. The entire IQDCA process can be expressed as follows:
\begin{equation}
    \operatorname{IQDCA}(\mathbf{Q}_P,\mathbf{K}_{ID},\mathbf{V}_{ID}) = \mathbf{A}\cdot \mathbf{V}_{ID}.
\end{equation}
The output of the $\operatorname{IQDCA}$ is $\mathbf{P}_{q} \in \mathbb{R}^{n \times N\times d}$, which contains all the weighted invariant queries. Then, we apply a self-weighted attention layer to discard irrelevant tokens to obtain the final affordance dictionary-based adaptive fusion feature $\mathbf{P}_{mix}$:
\begin{equation}
    \mathbf{P}_{mix} = \operatorname{SWA}([\mathbf{P}_{q_1}, \mathbf{P}_{q_2},\dots, \mathbf{P}_{q_n}]).
\end{equation}
Next, this adaptively weighted fusion feature is combined with the original point cloud feature $\mathbf{P}_{in}$ to obtain the final feature $\mathbf{P}_{out}$, which is then fed into the affordance decoder to predict the distribution of the affordance region.

\subsection{Decoder and Loss Function}
\label{Section:3.4}
We pool the last layer outputs of the IAM, $\mathbf{F}_1, \mathbf{F}_2,\dots, \mathbf{F}_n$, to compute the affordance logits $\hat{y}$. Additionally, we send $\mathbf{P}_{out}$ to the decoder $f_d$ to ground the 3D affordance $\mathbf{P}_{pred}$:
\begin{equation}
    \mathbf{P}_{pred} = f_{d}(\mathbf{P}_{out}),
\end{equation}
where $ \mathbf{P}_{pred} \in\mathbb{R}^{N \times 1}$. The total loss consists of three components: $\mathbf{\mathcal{L}}_{CE}$, $ \mathbf{\mathcal{L}}_{Sim}$ and $\mathbf{\mathcal{L}}_{HM}$. $\mathbf{\mathcal{L}}_{CE}$ represents the cross-entropy loss between $y$ and $\hat{y}$. $\mathbf{\mathcal{L}}_{Sim}$ is the cosine similarity loss calculated at each layer of $\{\mathbf{F}_1, \mathbf{F}_2,\dots, \mathbf{F}_n\}$, which aims to align image features while obtaining invariant affordance knowledge. $\mathbf{\mathcal{L}}_{HM}$ combines focal loss~\cite{lin2017focal} with dice loss~\cite{milletari2016v}, and is calculated between $\mathbf{P}_{pred}$ and $\mathbf{P}_{label}$, supervising the point-wise heatmap on point clouds. The total loss is formulated as follows:
\begin{equation}
\label{Equ:loss}
    \mathbf{\mathcal{L}}_{total} = \lambda_{1}\mathbf{\mathcal{L}}_{CE} + \lambda_{2}\mathbf{\mathcal{L}}_{Sim} + \lambda_{3}\mathbf{\mathcal{L}}_{HM},
\end{equation}
where $\lambda_1$, $\lambda_2$, and $\lambda_3$ are hyperparameters used to balance the total loss. Further details can be found in the supplementary materials.

\section{Experiments}
\label{sec:exp}

\subsection{Experimental Settings}

\subsubsection{Dataset}
To the best of our knowledge, there is currently no affordance dataset that satisfactorily tackles challenges posed by significant variations in appearance. Common datasets lack visual information or primarily focus on a single object corresponding image, ignoring the invariant affordance knowledge and internal relationships across different object contexts. To address these challenges, we constructed the \textbf{M}ulti-\textbf{I}mage and \textbf{P}oint \textbf{A}ffordance (\textbf{\dataset}) Dataset, which comprised paired multi-images and point clouds. We leverage point clouds and affordance annotations from 3D AffordanceNet~\cite{deng20213d}, while gathering paired multiple images from IAGNet~\cite{yang2023grounding}, HICO~\cite{chao2015hico} and AGD20K~\cite{luo2022learning}. The proposed \textbf{\dataset} dataset contains 5,162 images and 7,012 point clouds, covering 23 object classes and 17 affordance categories. In addition, we conducted our training and evaluation under the \textbf{\texttt{seen}} and \textbf{\texttt{unseen}} settings following previous works~\cite{yang2023grounding, li2024laso}. The \textbf{\texttt{seen}} setting shares identical object categories in training and evaluation, whereas the \textbf{\texttt{unseen}} setting utilizes different splits of objects.

\subsubsection{Compared Baselines and Evaluation Metrics}
The most relevant works are IAGNet~\cite{yang2023grounding} and LASO~\cite{li2024laso}, both of which leverage 3D AffordanceNet~\cite{deng20213d} to derive the corresponding 3D shapes. In addition, consistent with IAGNet, we adopt SOTA image-3D cross-modal methods as baselines for a more comprehensive evaluation, encompassing PMF~\cite{zhuang2021perception}, XMF~\cite{aiello2022cross}, and ILN~\cite{ILN}, which concentrate on fusing image and point features, as well as MBDF~\cite{tan2021mbdf}, FRCNN~\cite{xu2022fusionrcnn}, and PFusion~\cite{xu2018pointfusion} which are predominantly employed for multi-modal object grounding. We reproduce these methods with the same feature extractors and settings in the original papers to ensure a fair comparison. Besides, we enhance these methods by replacing the single image or text inputs with multiple ones, allowing for a more direct comparison with our dataset. All the evaluation metrics follow previous works:
\textbf{A}rea \textbf{U}nder the \textbf{C}urve (\textbf{AUC})~\cite{lobo2008auc}, \textbf{a}verage \textbf{I}ntersection \textbf{O}ver \textbf{U}nion (\textbf{aIOU})~\cite{rahman2016optimizing}, \textbf{SIM}ilarity (\textbf{SIM})~\cite{swain1991color} and \textbf{M}ean \textbf{A}bsolute \textbf{E}rror (\textbf{MAE})~\cite{willmott2005advantages}.

\subsubsection{Implementation Details}
Following IAGNet~\cite{yang2023grounding}, we employ PointNet++~\cite{qi2017pointnet++} and ResNet-18~\cite{he2016deep} as the default 3D and 2D backbone, respectively. We train \ours model on a single NVIDIA A100 GPU with a batch size of 64, using the Adam optimizer with a learning rate of 4e-5. More detailed experiment settings can be found in the supplementary materials.

\subsection{Quantitative Analysis}
\cref{tab:main_results} reports the metrics of the proposed method \textbf{\ours} compared with other methods on the \textbf{\dataset benchmark}. The results demonstrate that the proposed \textbf{\ours} achieves state-of-the-art performance, significantly surpassing existing methods in 3D point cloud affordance prediction. Specifically, \textbf{\ours} achieve scores of 85.10 in \textbf{AUC} and 20.50 in \textbf{aIOU}, outperforming the second-best method, LASO~\cite{li2024laso}, and IAGNet~\cite{yang2023grounding}, with an improvement of +1.97 in \textbf{AUC} and +2.58 in \textbf{aIOU}. Notably, our method also demonstrates impressive affordance capabilities in unseen point cloud prediction, outperforming the second-best method, IAGNet by 1.22 in \textbf{AUC} and LASO by 0.05 in \textbf{aIOU}. Additionally, 3D segmentation and tracking methods such as PMF~\cite{zhuang2021perception}, XMF~\cite{aiello2022cross}, and MBDF~\cite{tan2021mbdf} perform significantly worse compared to the affordance method.

\subsection{Qualitative Results}

\begin{figure*}
    \centering
    \includegraphics[width=\textwidth]{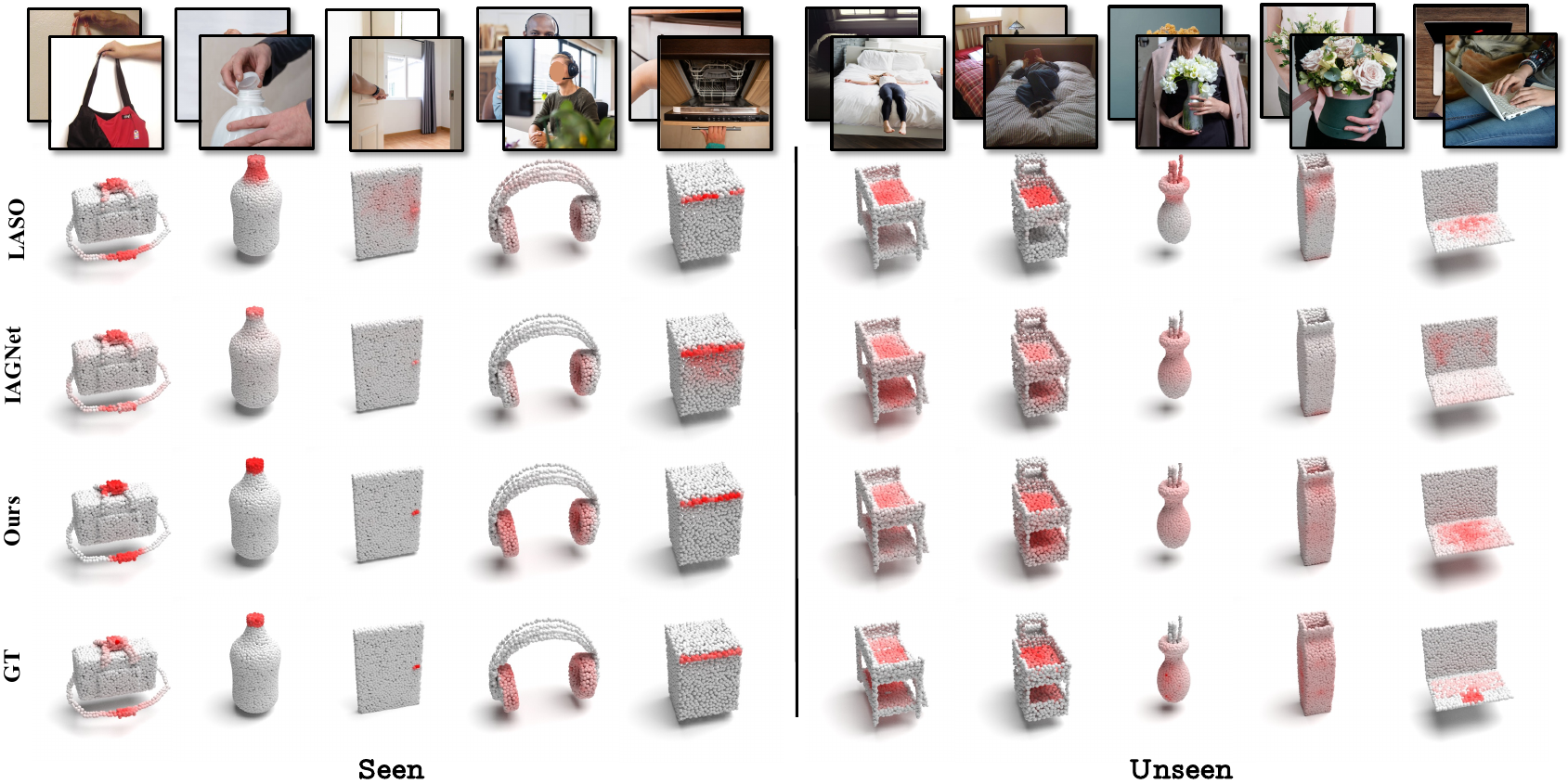}
    \caption{\textbf{Affordance Visualization on \textbf{\dataset} dataset.} Compared with LASO~\cite{li2024laso} and IAGNet~\cite{yang2023grounding}, the proposed \ours achieves more accurate results in both seen and unseen settings.}
    \label{fig:vis_res_1}
\end{figure*}

\subsubsection{Visualization of point cloud affordance grounding.}
\cref{fig:vis_res_1} illustrates the point cloud affordance grounding results of various methods in both \textbf{\texttt{seen}} and \textbf{\texttt{unseen}} scenes of the \textbf{\dataset} dataset. The results demonstrate that our method achieves more accurate results, outperforming LASO and IAGNet. For instance, on objects such as the ``door'' and the ``vase'', LASO and IAGNet suffer from affordance map dispersion or omission, whereas our approach can accurately focus on the interactive regions. The results also show that LASO and IAGNet, which ignore consistent cross-image features and rely solely on raw images, can lead to inaccurate area predictions.
\begin{figure}[t]
    \centering
    \includegraphics[width=\columnwidth]{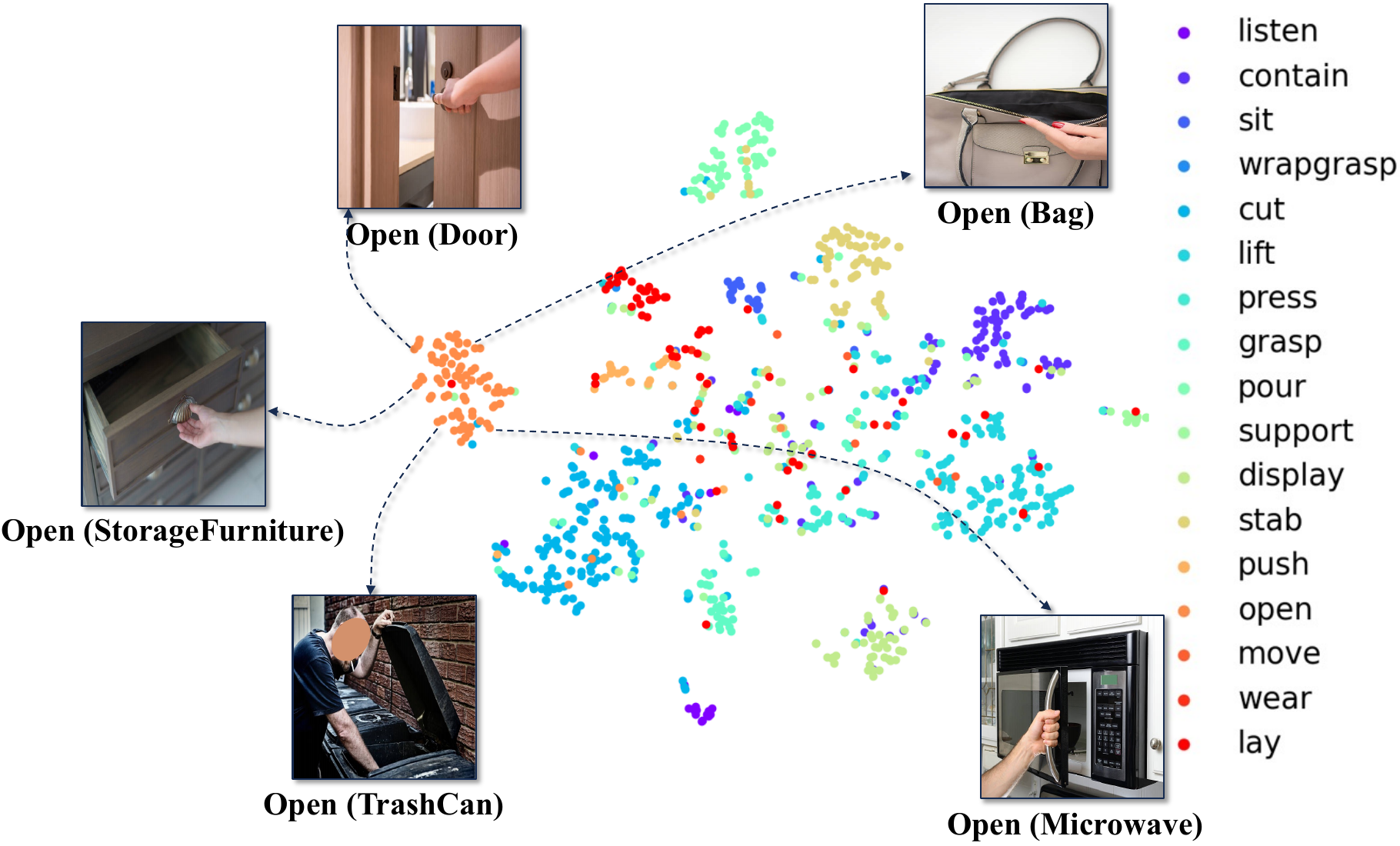}
    \caption{\textbf{t-SNE visualization of affordance queries.} Tokens query corresponding to the same operation across different object clusters in the region.}
    \label{fig:tsne}
    \vspace{-1.2em}
\end{figure}

\subsubsection{Qualitative Analysis by t-SNE.} 
In~\cref{fig:tsne}, we visualize the distribution of query tokens as they propagate through the IAM network, demonstrating its effectiveness in extracting invariant affordance knowledge. The affordance query tokens are randomly initialized at the beginning. As the affordance query tokens iteratively propagate, they gradually cluster based on affordance type, as shown in~\cref{fig:tsne}. For instance, features corresponding to the same ``open'' operation across different objects, such as ``Door'', ``Microwave'', and ``Bag'', cluster in the orange region. This result demonstrates the effectiveness of our approach by precisely extracting invariant knowledge across images.

\subsection{Ablation Study}

\begin{table}
\centering
\small

\resizebox{\columnwidth}{!}{
\begin{tabular}{ccc|cccc}
\toprule
 & \textbf{IAM} & \textbf{ADM} &\textbf{AUC $\uparrow$} & \textbf{aIOU $\uparrow$} & \textbf{SIM $\uparrow$} & \textbf{MAE $\downarrow$} \\ 
\midrule
\multirow{4}{*}{\rotatebox{90}{\textbf{\texttt{Seen}}}} &  &  &  82.97 & 16.88 & 0.519 & 0.107 \\
& \checkmark &  &  84.80 & 20.28 & 0.555 & 0.092 \\
 & & \checkmark & 84.57 & 17.35 & 0.536 & 0.117 \\
 & \checkmark & \checkmark & \textbf{85.10} & \textbf{20.50} & \textbf{0.568} & \textbf{0.091} \\
 
 \midrule
 
\multirow{4}{*}{\rotatebox{90}{\textbf{\texttt{Unseen}}}} & &  & 65.14 & 4.62 & 0.311 & 0.145 \\
& \checkmark &  & 70.34 & 4.96 & 0.312 & 0.150 \\
 & & \checkmark & 69.91 & 5.21 & 0.300 & 0.141 \\
 & \checkmark & \checkmark & \textbf{71.13} & \textbf{5.23} & \textbf{0.315} & \textbf{0.136} \\
 \bottomrule
\end{tabular}
}
\caption{\textbf{Ablation of IAM and ADM.} We investigate the improvement of IAM and ADM on the model performance based on the baseline.}
\label{tab:ablation}
\vspace{-1.2em}
\end{table}
\subsubsection{Effectiveness of IAM and ADM.} \cref{tab:ablation} reports the effect of IAM and ADM in both \textbf{\texttt{seen}} and \textbf{\texttt{unseen}} test datasets. Our full model outperforms in all metrics, such as higher \textbf{AUC} scores of 85.10 and 71.13 than the model w/o ADM (84.80 and 70.34) and AIM (82.97 and 65.14) on the \textbf{\texttt{Seen}} and \textbf{\texttt{Unseen}} setting, respectively. Notably, the IAM module leads to significant performance, while the ADM module has a smaller impact cause ADM depends on the IAM and needs the assistance of aligned features. Thus, when IAM and ADM are used together, will get the best performance across all metrics like the last row of Table~\ref{tab:ablation} in seen and unseen. Notably, the IAM module significantly enhances performance, while the ADM individual module has a smaller impact due to its reliance on IAM and feature alignment. Consequently, the best results across all metrics, both seen and unseen, are achieved when IAM and ADM are combined, as shown in the last row of~\cref{tab:ablation}.

\subsubsection{Effect of the image numbers.} \cref{tab:num_images} show the performance of \ours with frame input number from 1 to 5. The experiment results indicate a notable performance enhancement in a seen scene with an increasing image input, as evidenced by improvements in \textbf{AUC} (from 83.33 to 85.47) and \textbf{aIOU} (from 19.32 to 20.35) metrics. Nevertheless, an overabundance of image input can lead to a decline in affordance prediction performance for occluded point clouds, as reflected in the decrease in \textbf{aIOU} from 4.72 to 4.49. It indicates that a limited number of multi-images with significant variations may mutually interfere with the alignment process. In contrast, a sufficient number of images enables the model to distill generalizable, invariant features, resulting in enhanced performance.

\begin{table}[t]
\centering
\small
\resizebox{\columnwidth}{!}{
\begin{tabular}{c r c c c c c}
\toprule
\multirow{2}{*}{} & \multirow{2}{*}{\textbf{Metrics}}  & \multicolumn{5}{c}{\textbf{Number of Images}} \\ \cmidrule(lr){3-7} &  & 1 & 2 & 3 & 4 & 5 \\

\midrule
\multirow{4}{*}{\rotatebox{90}{\textbf{\texttt{Seen}}}} 
&  AUC $\uparrow$  & 83.33 & 84.15 & 83.68 & 83.78 & 85.47 \\
&  aIOU $\uparrow$  & 19.32 & 20.16 & 20.08 & 18.71 & 20.35 \\ 
&  SIM $\uparrow$  & 0.533 & 0.550 & 0.556 & 0.550 & 0.559 \\ 
&  MAE $\downarrow$  & 0.093 & 0.093 & 0.094 & 0.094 & 0.090 \\ 

\midrule

\multirow{4}{*}{\rotatebox{90}{\textbf{\texttt{Unseen}}}} 
&  AUC $\uparrow$  & 71.05 & 70.00 & 70.68 & 70.63 & 71.32  \\ 
&  aIOU $\uparrow$  & 4.72 & 4.15 & 5.15 & 4.15 & 4.49  \\ 
&  SIM $\uparrow$  & 0.314 & 0.294 & 0.301 & 0.318 & 0.318  \\ 
&  MAE $\downarrow$  & 0.146 & 0.151 & 0.158 & 0.154 & 0.128  \\ 

\bottomrule
\end{tabular}
}
\caption{\textbf{Ablation of the image numbers} on the \textbf{\dataset} dataset.}
\label{tab:num_images}
\vspace{-1.2em}
\end{table}

\subsubsection{Ablation of iterative layer numbers.} Table~\ref{tab:num_layers} evaluates the invariant affordance feature extraction capability of IAM with different iterative layers. Intuitively, increasing the iterative layer numbers can enhance the ability to extract invariant information. However, the best layer settings for seen and unseen scenarios are four and six, achieving the highest AUC (85.10 and 71.13) and SIM (0.568 and 0.315), respectively. This discrepancy arises from a sharp increase in similarity loss across excessive layers, which can cause a sharp increase in loss, leading to training instability and performance degradation.

\begin{table}[ht]
\centering
\small
\resizebox{\columnwidth}{!}{
\begin{tabular}{c r c c c c c c}
\toprule
\multirow{2}{*}{} & \multirow{2}{*}{\textbf{Metrics}}  & \multicolumn{6}{c}{\textbf{Number of Layers}} \\ \cmidrule(lr){3-8} &  & 1 & 2 & 3 & 4 & 5 & 6 \\

\midrule
\multirow{4}{*}{\rotatebox{90}{\textbf{\texttt{Seen}}}} 
&  AUC $\uparrow$  & 84.42 & 84.35 & 84.56 & 85.10 & 83.92 & 83.37 \\
&  aIOU $\uparrow$  & 19.83 & 20.44 & 20.70 & 20.50 & 19.32 & 19.56 \\ 
&  SIM $\uparrow$  & 0.556 & 0.556 & 0.560 & 0.568 & 0.545 & 0.554 \\ 
&  MAE $\downarrow$  & 0.093 & 0.093 & 0.092 & 0.091 & 0.099 & 0.094 \\ 

\midrule

\multirow{4}{*}{\rotatebox{90}{\textbf{\texttt{Unseen}}}} 
&  AUC $\uparrow$  & 70.60 & 69.06 & 70.87 & 70.30 & 70.94 & 71.13 \\ 
&  aIOU $\uparrow$  & 4.70 & 4.47 & 4.45 & 4.72 & 4.93 & 5.23 \\ 
&  SIM $\uparrow$  & 0.309 & 0.313 & 0.313 & 0.311 & 0.309 & 0.315 \\ 
&  MAE $\downarrow$  & 0.131 & 0.160 & 0.164 & 0.143 & 0.153 & 0.136 \\ 

\bottomrule
\end{tabular}
}
\caption{\textbf{Ablation of iterative layer numbers} on the \textbf{\dataset} dataset.}
\label{tab:num_layers}
\vspace{-1.2em}
\end{table}

\subsection{Real-World Evaluation}
\begin{figure}[t!]
    \centering
    \includegraphics[width=\columnwidth]{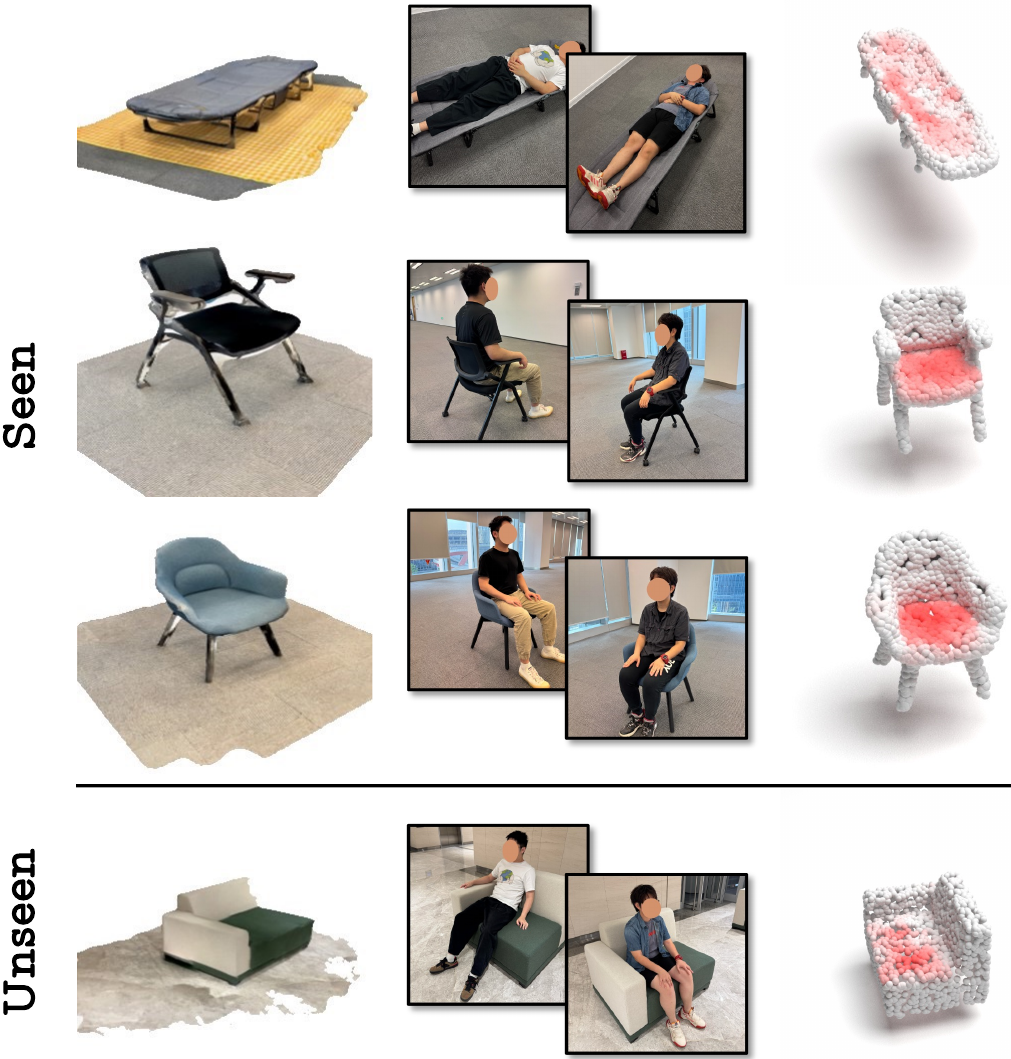}
    \caption{\textbf{Real-World Visualization.} \textbf{Left:} Original 3D point clouds scanned by an iPhone 15 Pro. \textbf{Middle:} Reference images. \textbf{Right:} Affordance prediction results on the scanned point cloud.}
    \label{fig:real_world}
    \vspace{-1.2em}
\end{figure}

To evaluate the zero-shot generalization ability of our method in real-world scenarios, we test it on real scenes, as shown in~\cref{fig:real_world}.

Specifically, we use an iPhone 15 Pro equipped with LiDAR to scan real-world objects and generate their point clouds. These point clouds, along with captured human-object interaction reference images, are then fed into our trained model to generate affordance predictions on the point clouds. We conduct the real-world evaluation under two settings: \textbf{\texttt{seen}} and \textbf{\texttt{unseen}}, in a manner consistent with previous experiments. The ``bed'' and the ``chair'' are objects that are present in the dataset, while the ``sofa'' is not included in the training dataset. Notably, objects in both settings are new to our trained model, as their point clouds are built from scratch by ourselves. The robust visualization results demonstrate the effective generalization of our method.

\section{Conclusion and Limitations}

In this work, we propose the \textbf{M}ulti-\textbf{I}mage Guided Invariant-\textbf{F}eature-Aware 3D \textbf{A}ffordance \textbf{G}rounding (\textbf{\ours}) framework. Our approach gradually extracts affordance knowledge from multiple human-object reference images and effectively integrates this invariant knowledge with point cloud representations to achieve accurate affordance prediction. Moreover, we construct the \textbf{M}ulti-\textbf{I}mage and \textbf{P}oint \textbf{A}ffordance (\textbf{\dataset}) benchmark to advance research in understanding the affordances of 3D objects. Extensive experiments are conducted on the \dataset dataset and our method outperforms previous state-of-the-art methods.

\subsubsection{Limitations} Despite the demonstrated effectiveness of \ours, it is important to acknowledge certain limitations. Specifically, the 3D objects in our dataset are still relatively simple compared to real-world scenarios. Moreover, our work primarily focuses on visual understanding of affordances, without accounting for manipulation, \eg, the size of the manipulator, the direction of the manipulation action, \etc In future work, we aim to bring our approach closer to real-world embodied manipulation. Further discussions and analysis are provided in the supplementary materials.

\clearpage
\bibliography{aaai25}

\clearpage
\appendix
\section{Dataset}
\label{apdix:sec:dataset}
Reference human-object interaction images that indicate the same affordance of objects often vary significantly in appearance, yet they all reveal the same affordance category and thus have strong internal relationships and inherent consistency. To leverage the invariant affordance knowledge across multiple images, we construct the Multi-Image and Point Affordance (\textbf{\dataset}), which for the first time combines the point clouds with multiple reference images.

The point clouds and affordance annotations in \dataset are sourced from 3D AffordanceNet~\cite{deng20213d}, and the reference images are from IAGNet~\cite{yang2023grounding}, HICO~\cite{chao2015hico} and AGD20K~\cite{luo2022learning}. The dataset contains 7,012 point clouds and 5,162 images covering 23 object classes and 17 affordance categories. Consistent with IAGNet, we construct two types of dataset with the \textbf{\texttt{seen}} and \textbf{\texttt{unseen}} settings. Finally, in the training split, \dataset includes 6,000 point clouds, each paired with multiple images from each corresponding affordance type. In the test split, \dataset includes 1,000 point clouds, each paired with multiple images from a specific affordance type.

\section{Additional Implementation Details}
\label{apdix:sec:imple}

\subsection{Network Details}

Following previous efforts~\cite{yang2023grounding,li2024laso}, we employ ResNet-18 and PointNet++ as the encoders for images and point clouds, respectively. Specifically, for the image input, we first resize the images to $224\times224$ pixels and then apply random cropping. We do not rely on the bounding boxes of the subject and object introduced by IAGNet. For the point cloud input, we use a 3-layer Set Abstraction module to extract point cloud features, resulting in feature dimensions of (320, 512), (512, 128), and (512, 64), where 320, 512, and 512 represent the hidden layer dimensions, and 512, 128, and 64 represent the number of abstracted points. Each set abstraction layer uses the farthest point sampling strategy.

\subsection{Training Details}
We train our model and implement all baseline methods under the same training settings for a fair comparison. Specifically, we randomly sample multiple images from each affordance type during training. We use PyTorch to implement our methods and all baseline methods. The Adam~\cite{kingma2017adammethodstochasticoptimization} optimizer is employed with betas set to 0.9 and 0.999. The initial learning rate is set to 4e-5, and a cosine learning rate scheduler is used with $\text{T}_{\text{max}}$ set to the total number of epochs. Each model is trained on a single NVIDIA A100 GPU with a batch size of 64 for a total of 80 epochs.

\subsection{Loss Functions}
We provide detailed explanations of the loss functions used in \ours. The total loss is formulated as follows:
\begin{equation}
     \mathbf{\mathcal{L}}_{total} = \lambda_{1}\mathbf{\mathcal{L}}_{CE} + \lambda_{2}\mathbf{\mathcal{L}}_{Sim} + \lambda_{3}\mathbf{\mathcal{L}}_{HM}.
\end{equation}

\subsubsection{Cross-entropy Loss}
We use the cross entropy loss $\mathcal{L}_{CE}$ to supervise the predicted affordance category with the ground truth category, formulated as follows:
\begin{equation}
   \mathcal{L}_{CE} = - \sum_{i=1}^K y_i\log(\hat{y}_i),
\end{equation}
where $K$ denotes the number of affordance categories, which is 17 in our method.

\subsubsection{Similarity Loss} 
To enable our \ours to extract consistent information from images that share the same affordance type but exhibit significant appearance variations, we compute the cosine similarity loss $\mathcal{L}_{Sim}$ between multiple images at the output of each iterative layer in the image feature branch of the IAM:
\begin{equation}
    \scalebox{0.93}{$
    \mathcal{L}_{Sim} = \frac{1}{L} \sum_{l=1}^{L}(1-\frac{\mathbf{F}_i^{(l)} \cdot \mathbf{F}_j^{(l)}}{\| \mathbf{F}_i^{(l)}  \| \| \mathbf{F}_j^{(l)}  \|}), \forall {i,j} \in \{ 1,2,\dots,n\}, i\neq j
    $},
\end{equation}
where $L$ denotes the number of iterative layers, $\mathbf{F}_i^{(l)}$ and $\mathbf{F}_j^{(l)}$ denotes the features of reference image $i$ and $j$ from layer $l$, respectively.

\subsubsection{Heatmap Loss} 
Consistent with IAGNet~\cite{yang2023grounding}, the heatmap loss $\mathcal{L}_{HM}$ consists of a focal loss~\cite{lin2017focal} and a dice loss~\cite{milletari2016v}, which are used to supervise the predicted affordance heatmap and the corresponding ground truth. This can be formulated as follows:
\begin{align}
    \mathcal{L}_{HM} &= \mathcal{L}_{\text{focal}} + \mathcal{L}_{\text{dice}},\\
    \mathcal{L}_{\text{focal}} &= -\alpha \sum_{i=1}^{N} ( (1-\hat{p}_i)^\gamma p_i \log(\hat{p}_i) \nonumber \\ &\quad +(1-\alpha)(\hat{p}_i^\gamma (1-p_i) \log(1-\hat{p}_i))),\\
    \mathcal{L}_{\text{dice}} &= 1 - \frac{2 \sum_{i=1}^{N} \hat{p}_i p_i + \epsilon}{\sum_{i=1}^{N} \hat{p}_i + \sum_{i=1}^{N} p_i + \epsilon},
\end{align}
where $\alpha$ and $\gamma$ are hyperparameters of the focal loss, set to 0.25 and 2 respectively in our implementation. $\hat{p}_i$ and $p_i$ represent the predicted affordance heatmap and the corresponding ground truth, respectively, and $\epsilon$ is a small constant used to prevent division by zero.

\subsection{Evaluation Metrics}
Here, we provide a detailed explanation of the evaluation metrics used in our study. Following previous research, we adopt AUC~\cite{lobo2008auc}, aIOU~\cite{rahman2016optimizing}, SIM~\cite{swain1991color}, and MAE~\cite{willmott2005advantages} as our evaluation metrics:

\begin{itemize}
    \item \textbf{AUC: Area Under the Curve.} AUC refers to the area under the ROC curve. The ROC curve is generated by plotting the relationship between the False Positive Rate (FPR) and the True Positive Rate (TPR). The value of AUC ranges from 0 to 1, with values closer to 1 indicating better classification performance. In our evaluation, the model's predictions are treated as 2048 binary classification results (corresponding to the number of points in the point cloud), formulated as follows:
    \begin{align}
        \text{TPR} &= \frac{\text{TP}}{\text{TP}+\text{FN}},\\
        \text{FPR} &= \frac{\text{FP}}{\text{FP}+\text{TN}}.
    \end{align}

    \item \textbf{aIOU: average Intersection Over Union.} aIOU is a metric used to measure the overlap between the predicted region and the ground truth region. The formula is as follows:
    \begin{equation}
        \text{IOU}=\frac{\text{Intersction Area}}{\text{Union Area}}.
    \end{equation}
    We binarize the prediction results using multiple thresholds ranging from 0 to 0.99, with an interval of 0.01. The aIOU is the average IOU calculated across these thresholds for multiple samples:
    \begin{equation}
        \text{aIOU}=\frac{1}{T}\sum_{i=1}^{T}\text{IOU}_i,
    \end{equation}
    where $T$ denotes the number of thresholds.

    \item \textbf{SIM: Similarity.} SIM is used to measure the similarity between the predicted map \(P\) and the continuous ground truth map \(Q_D\), which can be formulated as follows:
    \begin{equation}
        \text{SIM}(P, Q_D) = \sum_{i=1}^{N} \min\left(\frac{P_i}{\max(P)}, \frac{(Q_D)_i}{\max(Q_D)}\right),
    \end{equation}
    where \(N\) is the number of points, \(P_i\) represents the predicted value for the \(i\)th point, and \((Q_D)_i\) represents the corresponding ground truth value.

    \item \textbf{MAE: Mean Absolute Error.} MAE represents the Mean Absolute Error between the predicted values and the actual values:
    \begin{equation}
    \text{MAE}=\frac{1}{N}\sum_{i=1}^{N}|y_i-\hat{y}_i|,
    \end{equation}
    where \(N\) is the number of points, \(y_i\) is the ground truth value for the \(i\)th point, and \(\hat{y_i}\) is the predicted value for the \(i\)th point.

\end{itemize}

\subsection{Compared Baselines}
The works most relevant to ours are IAGNet~\cite{yang2023grounding} and LASO~\cite{li2024laso}, both of which leverage 3D AffordanceNet~\cite{deng20213d} to obtain the corresponding 3D shapes and focus on 3D affordance grounding. We use the official codes of these methods to conduct comparative experiments. Moreover, consistent with IAGNet, we adopt several SOTA methods in image-3D cross-modal understanding as baselines for a more comprehensive evaluation. We reimplement all these methods' network architecture and conduct training and evaluation on our \dataset.

\subsubsection{PMF~\cite{zhuang2021perception}}
PMF introduces a Perception-Aware Multi-Sensor Fusion scheme for 3D LiDAR semantic segmentation. This method employs a two-stream network with residual-based fusion modules and introduces perception-aware losses to measure and integrate the perceptual differences between the two data modalities.

\subsubsection{MBDF~\cite{tan2021mbdf}}
MBDF introduces the Multi-Branch Deep Fusion Network for 3D object detection, which fuses point cloud and image data to improve detection accuracy. MBDF utilizes Adaptive Attention Fusion (AAF) modules to integrate single-modal semantic features and generate cross-modal fusion features, employs a region of interest (RoI)-pooled fusion module to refine proposals, and introduces a hybrid sampling strategy for point cloud downsampling.

\subsubsection{FRCNN~\cite{xu2022fusionrcnn}}
FRCNN introduces a two-stage LiDAR-camera fusion approach for 3D object detection, integrating sparse geometric information from LiDAR point clouds with dense textural information from camera images using a unified attention mechanism. The method addresses the challenge of accurately recognizing and locating objects, particularly those with sparse point clouds at greater distances, by enhancing domain-specific features through intra-modality self-attention and fusing the modalities via cross-attention.

\subsubsection{ILN~\cite{ILN}}
ILN is designed for the registration of low-overlap point cloud pairs with the assistance of misaligned intermediate images. ILN extracts deep features from three modalities, fuses them using attention modules, and employs a two-stage classifier to progressively identify overlapping regions, ultimately establishing soft correspondences for accurate registration.

\subsubsection{PFusion~\cite{xu2018pointfusion}}
PFusion presents a 3D object detection method that integrates image and 3D point cloud data to estimate 3D bounding boxes of objects. It processes image data through a CNN and raw point cloud data through a PointNet architecture, then combines these using a fusion network that predicts multiple 3D box hypotheses relative to input 3D points as spatial anchors.

\subsubsection{XMF~\cite{aiello2022cross}}
XMF introduces an architecture for point cloud completion guided by an auxiliary image, addressing the challenge of fusing multimodal data from point clouds and images to complete 3D shapes. The method leverages cross-attention operations in the latent space to integrate localized features from both modalities, avoiding complex reconstruction techniques and offering a flexible decoder for diverse completion tasks.

\subsubsection{IAGNet~\cite{yang2023grounding}}
IAGNet introduces a framework for grounding 3D object affordances from 2D interactions in images. It presents the Point-Image Affordance Dataset (PIAD) to support this task and addresses the challenge of aligning object region features across different modalities.

\onecolumn
\stepcounter{table}
\begin{longtable*}{cr ccccccccc}
\toprule
\textbf{Affordances} & \textbf{Metrics} & \textbf{MBDF} & \textbf{PMF} & \textbf{FRCNN} & \textbf{ILN} & \textbf{PFusion} & \textbf{XMF} & \textbf{IAGNet} & \textbf{LASO} & \textbf{\ours} \\ \midrule
\endfirsthead

\multicolumn{11}{r}{{Continued from previous page}} \\
\toprule
\textbf{Affordances} & \textbf{Metrics} & \textbf{MBDF} & \textbf{PMF} & \textbf{FRCNN} & \textbf{ILN} & \textbf{PFusion} & \textbf{XMF} & \textbf{IAGNet} & \textbf{LASO} & \textbf{\ours} \\ 
\midrule
\endhead

\multicolumn{11}{r}{{Continued on next page}} \\ 
\endfoot

\\[\dimexpr \belowcaptionskip -\normalbaselineskip]
\endlastfoot

\multirow{4}{*}{\textbf{grasp}}   & AUC $\uparrow$  & 67.93  & 72.74  & 71.10  & 72.10  & 70.78  & 70.14  & 79.25  & 71.57  & 72.18  \\
                                   & aIOU $\uparrow$ & 4.85   & 4.61   & 4.32   & 4.48   & 4.55   & 4.27  & 13.93  & 8.17  & 13.52  \\
                                   & SIM $\uparrow$  & 0.403  & 0.396  & 0.401  & 0.394  & 0.400  & 0.385  & 0.514  & 0.491  & 0.518  \\
                                   & MAE $\downarrow$  & 0.150  & 0.153  & 0.154  & 0.155  & 0.155  & 0.159  & 0.100  & 0.118  & 0.110  \\ \midrule
\multirow{4}{*}{\textbf{contain}}    & AUC $\uparrow$ & 83.60  & 83.38  & 83.31  & 83.49  & 81.96  & 83.82  & 78.07  & 82.15  & 81.06  \\
                                   & aIOU $\uparrow$ & 12.46   & 9.49   & 10.13   & 9.26  & 9.31   & 9.03  & 14.31  & 12.25  & 15.41  \\
                                   & SIM $\uparrow$ & 0.489  & 0.477  & 0.484  & 0.469    & 0.455  & 0.474  & 0.514  & 0.489  & 0.533  \\
                                   & MAE $\downarrow$ & 0.110  & 0.117  & 0.114  & 0.120  & 0.127  & 0.119  & 0.103  & 0.107  & 0.104  \\ \midrule
\multirow{4}{*}{\textbf{lift}}     & AUC $\uparrow$ & 51.06  & 92.06  & 78.98  & 88.41  & 71.14  & 90.92  & 92.12  & 96.05  & 87.62  \\
                                   & aIOU $\uparrow$ & 0.61  & 15.21  & 10.16  & 11.04  & 9.92  & 11.02  & 22.48  & 36.64  & 38.13  \\
                                   & SIM $\uparrow$  & 0.034  & 0.129  & 0.098  & 0.105  & 0.089  & 0.109  & 0.213  & 0.336  & 0.343  \\
                                   & MAE $\downarrow$ & 0.112  & 0.110  & 0.118  & 0.117  & 0.128  & 0.120  & 0.085  & 0.061  & 0.043  \\ \midrule
\multirow{4}{*}{\textbf{open}}     & AUC $\uparrow$  & 79.30  & 81.40  & 82.08  & 82.19  & 82.50  & 79.90  & 89.58  & 88.80  & 88.93  \\
                                   & aIOU $\uparrow$ & 6.69   & 7.73   & 7.60   & 7.98   & 8.69   & 7.49   & 22.09   & 20.33   & 23.85  \\
                                   & SIM $\uparrow$  & 0.197  & 0.195  & 0.196  & 0.198  & 0.209  & 0.193  & 0.346  & 0.364  & 0.372  \\
                                   & MAE $\downarrow$ & 0.103  & 0.114  & 0.113  & 0.115  & 0.123  & 0.119  & 0.052  & 0.061  & 0.046  \\ \midrule
\multirow{4}{*}{\textbf{lay}}     & AUC $\uparrow$  & 88.71  & 90.19  & 90.66  & 90.38  & 91.01  & 89.71  & 92.46  & 92.71  & 88.93  \\
                                   & aIOU $\uparrow$ & 21.68   & 17.35   & 19.65   & 17.78   & 19.40   & 17.67   & 26.57   & 9.17   & 23.85  \\
                                   & SIM $\uparrow$  & 0.594  & 0.594  & 0.610  & 0.598  & 0.614  & 0.593  & 0.659  & 0.672  & 0.372  \\
                                   & MAE $\downarrow$ & 0.104  & 0.108  & 0.103  & 0.106  & 0.105  & 0.109  & 0.091  & 0.090  & 0.046  \\ \midrule
\multirow{4}{*}{\textbf{sit}}   & AUC $\uparrow$ & 92.10  & 93.76  & 94.51  & 93.86  & 94.12  & 93.81  & 96.27  & 95.75  & 72.18  \\
                                   & aIOU $\uparrow$  & 29.59   & 18.97   & 18.59   & 8.02   & 19.64  & 18.55  & 36.36  & 35.59  & 13.52  \\
                                   & SIM $\uparrow$ & 0.646  & 0.523  & 0.521  & 0.393  & 0.532  & 0.515  & 0.724  & 0.712  & 0.518  \\
                                   & MAE $\downarrow$  & 0.085  & 0.119  & 0.121  & 0.144  & 0.119  & 0.123  & 0.065  & 0.068  & 0.110  \\ \midrule
\multirow{4}{*}{\textbf{support}}    & AUC $\uparrow$ & 80.35  & 77.07  & 75.83  & 77.40  & 78.94  & 78.13  & 85.68  & 84.21  & 81.06  \\
                                   & aIOU $\uparrow$ & 7.78   & 5.84   & 5.93   & 5.73   & 6.59   & 5.83  & 11.48  & 11.06  & 15.41  \\
                                   & SIM $\uparrow$ & 0.662  & 0.656 & 0.651  & 0.654  & 0.662  & 0.649  & 0.727  & 0.712  & 0.533  \\
                                   & MAE $\downarrow$ & 0.109  & 0.121  & 0.124  & 0.123  & 0.123  & 0.124  & 0.100  & 0.103  & 0.104  \\ \midrule
\multirow{4}{*}{\textbf{wrapgrasp}}     & AUC $\uparrow$ & 49.50  & 55.69  & 56.58  & 55.58  & 84.52  & 55.18  & 63.50  & 63.87  & 87.62  \\
                                   & aIOU $\uparrow$ & 2.82  & 2.65  & 3.30  & 2.62  & 11.43  & 2.61  & 3.914  & 4.35  & 38.13  \\
                                   & SIM $\uparrow$ & 0.518  & 0.514  & 0.527  & 0.512  & 0.223  & 0.517  & 0.596  & 0.558  & 0.343  \\
                                   & MAE $\downarrow$ & 0.152  & 0.153  & 0.149  & 0.155  & 0.133  & 0.153  & 0.131  & 0.140  & 0.043  \\ \midrule
\multirow{4}{*}{\textbf{pour}}     & AUC $\uparrow$ & 82.08  & 87.38  & 84.74  & 87.99  & 86.53  & 86.58  & 80.53  & 79.05  & 88.93  \\
                                   & aIOU $\uparrow$ & 5.21   & 9.85   & 9.47   & 10.32   & 10.04   & 9.69   & 11.81   & 11.32   & 23.85  \\
                                   & SIM $\uparrow$ & 0.364  & 0.408  & 0.397  & 0.410  & 0.407  & 0.402  & 0.462  & 0.412  & 0.372  \\
                                   & MAE $\downarrow$ & 0.147  & 0.130  & 0.131  & 0.131  & 0.132  & 0.134  & 0.103  & 0.117  & 0.046  \\ \midrule
\multirow{4}{*}{\textbf{move}}     & AUC $\uparrow$ & 57.63  & 60.75  & 59.82  & 60.35  & 60.19  & 60.04  & 61.88  & 59.54  & 88.93  \\
                                   & aIOU $\uparrow$ & 4.04   & 3.51   & 3.37   & 3.51   & 3.62   & 3.39   & 3.914   & 3.97   & 23.85  \\
                                   & SIM $\uparrow$ & 0.410  & 0.499  & 0.489  & 0.496  & 0.489  & 0.497  & 0.446  & 0.424  & 0.372  \\
                                   & MAE $\downarrow$ & 0.190  & 0.174  & 0.176  & 0.174  & 0.180  & 0.176  & 0.171  & 0.181  & 0.046  \\ \midrule
\multirow{4}{*}{\textbf{display}}   & AUC $\uparrow$ & 87.24  & 88.16  & 88.63  & 88.47  & 87.90  & 86.93  & 91.46  & 87.14  & 72.18  \\
                                   & aIOU $\uparrow$ & 22.76   & 21.23   & 22.40   & 22.96   & 24.45   & 19.94  & 29.01  & 26.37  & 13.52  \\
                                   & SIM $\uparrow$ & 0.588  & 0.595  & 0.602  & 0.608  & 0.603  & 0.577  & 0.674  & 0.631  & 0.518  \\
                                   & MAE $\downarrow$ & 0.104  & 0.103  & 0.101  & 0.100  & 0.111  & 0.110  & 0.083  & 0.092  & 0.110  \\ \midrule
\multirow{4}{*}{\textbf{push}}    & AUC $\uparrow$ & 69.50  & 76.14  & 76.73  & 78.54  & 69.90  & 78.67  & 77.63  & 84.57  & 81.06  \\
                                   & aIOU $\uparrow$ & 0.988   & 1.21   & 1.83   & 1.73   & 1.58   & 1.53  & 2.17  & 3.92  & 15.41  \\
                                   & SIM $\uparrow$  & 0.529  & 0.499  & 0.528  & 0.521  & 0.521  & 0.524  & 0.546  & 0.607  & 0.533  \\
                                   & MAE $\downarrow$ & 0.084  & 0.097  & 0.086  & 0.091  & 0.121  & 0.102  & 0.075  & 0.069  & 0.104  \\ \midrule
\pagebreak
\multirow{4}{*}{\textbf{listen}}     & AUC $\uparrow$ & 73.73  & 72.13  & 71.13  & 71.06  & 71.15  & 71.01  & 85.49  & 86.60  & 87.62  \\
                                   & aIOU $\uparrow$ & 8.60  & 4.41  & 4.33  & 4.31  & 4.28  & 4.60  & 11.25  & 13.42  & 38.13  \\
                                   & SIM $\uparrow$ & 0.500  & 0.405  & 0.410  & 0.405  & 0.402  & 0.417  & 0.594  & 0.638  & 0.343  \\
                                   & MAE $\downarrow$ & 0.132  & 0.170  & 0.169  & 0.170  & 0.168  & 0.168  & 0.109  & 0.100  & 0.043  \\ \midrule
\multirow{4}{*}{\textbf{wear}}     & AUC $\uparrow$  & 65.75  & 60.37  & 58.96  & 60.10  & 61.78  & 61.81  & 59.31  & 63.84  & 88.93  \\
                                   & aIOU $\uparrow$ & 2.31   & 2.77   & 2.50   & 2.77   & 2.56   & 2.82   & 2.59   & 3.78   & 23.85  \\
                                   & SIM $\uparrow$ & 0.560  & 0.526  & 0.528  & 0.530  & 0.535  & 0.538  & 0.545  & 0.506  & 0.372  \\
                                   & MAE $\downarrow$ & 0.144  & 0.163  & 0.166  & 0.164  & 0.164  & 0.164  & 0.146  & 0.159  & 0.046  \\ \midrule
\multirow{4}{*}{\textbf{press}}     & AUC $\uparrow$ & 86.79  & 80.49  & 82.11  & 81.39  & 84.14  & 82.20  & 87.93  & 88.93  & 88.93  \\
                                   & aIOU $\uparrow$ & 9.09   & 4.65   & 5.77   & 5.74   & 7.07   & 4.95   & 2.59   & 12.13   & 23.85  \\
                                   & SIM $\uparrow$ & 0.335  & 0.202  & 0.212  & 0.217  & 0.239  & 0.212  & 0.327  & 0.367  & 0.372  \\
                                   & MAE $\downarrow$ & 0.084  & 0.108  & 0.107  & 0.100  & 0.109  & 0.114  & 0.074  & 0.066  & 0.046  \\ \midrule
\multirow{4}{*}{\textbf{cut}}     & AUC $\uparrow$  & 89.69  & 89.42  & 88.56  & 89.19  & 86.44 & 90.61   & 63.16  & 87.55  & 88.93  \\
                                   & aIOU $\uparrow$ & 11.44   & 12.26   & 11.86   & 12.16   & 11.53   & 12.00   & 6.35   & 14.25   & 23.85  \\
                                   & SIM $\uparrow$ & 0.543  & 0.472  & 0.451  & 0.476  & 0.424  & 0.482  & 0.319  & 0.566  & 0.372  \\
                                   & MAE $\downarrow$ & 0.075  & 0.086  & 0.092  & 0.087  & 0.097  & 0.088  & 0.135  & 0.069  & 0.046  \\ \midrule
\multirow{4}{*}{\textbf{stab}}     & AUC $\uparrow$ & 93.41  & 99.88  & 99.85  & 99.96  & 99.86  & 99.86  & 33.02  & 78.03  & 88.93  \\
                                   & aIOU $\uparrow$ & 9.66   & 16.92   & 14.31   & 17.18   & 14.61   & 15.59   & 3.07   & 8.40   & 23.85  \\
                                   & SIM $\uparrow$  & 0.293  & 0.452  & 0.448  & 0.469  & 0.438  & 0.465  & 0.096  & 0.362  & 0.372  \\
                                   & MAE $\downarrow$ & 0.114  & 0.085  & 0.088  & 0.086  & 0.086  & 0.096  & 0.133  & 0.071  & 0.046  \\ \bottomrule

\caption*{Table \thetable: \textbf{Affordance prediction metrics of every affordance category on \dataset in the \texttt{seen} setting.}} \label{apdix:tab:res_1}
\end{longtable*}
\twocolumn

\subsubsection{LASO~\cite{li2024laso}}
LASO introduces the task of Language-guided Affordance Segmentation on 3D Objects, addressing the challenge of segmenting parts of 3D objects relevant to specific affordance questions posed in natural language. It proposes a dataset and a baseline model named PointRefer, which employs an Adaptive Fusion Module and a Referred Point Decoder to integrate textual cues with 3D object point clouds for semantic-aware segmentation.

\section{Additional Experimental Results}
\label{apdix:sec:exp}

\begin{figure*}[t]
    \centering
    \includegraphics[width=0.85\textwidth]{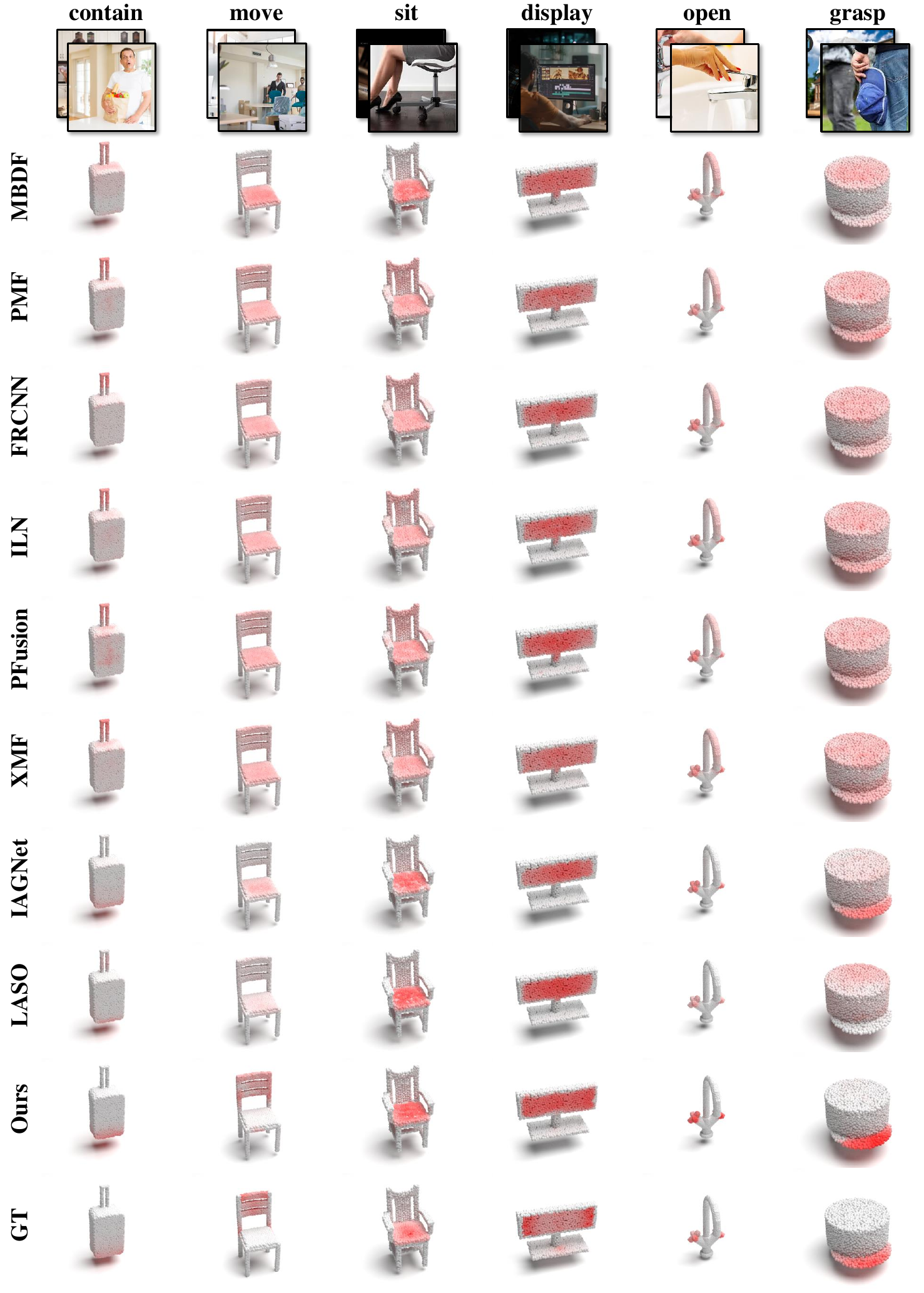}
    \caption{\textbf{Visualization results in setting \texttt{seen}.}}
    \label{apdix:fig:ap_vis1}
\end{figure*}

\begin{figure*}[t]
    \centering
    \includegraphics[width=0.71\textwidth]{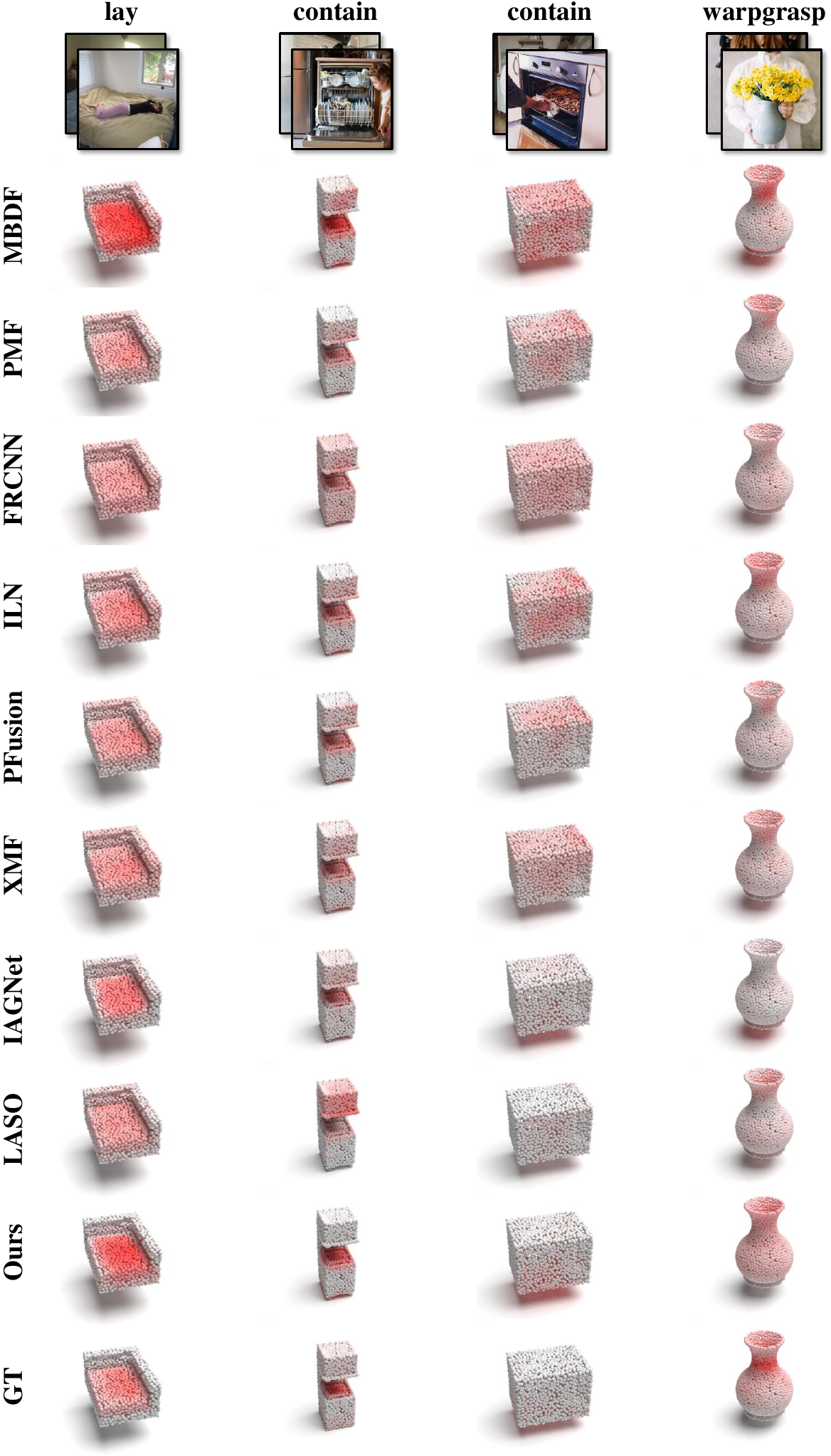}
    \caption{\textbf{Visualization results in setting \texttt{unseen}.}}
    \label{apdix:fig:ap_vis2}
\end{figure*}

\begin{figure*}[t]
    \centering
    \includegraphics[width=\textwidth]{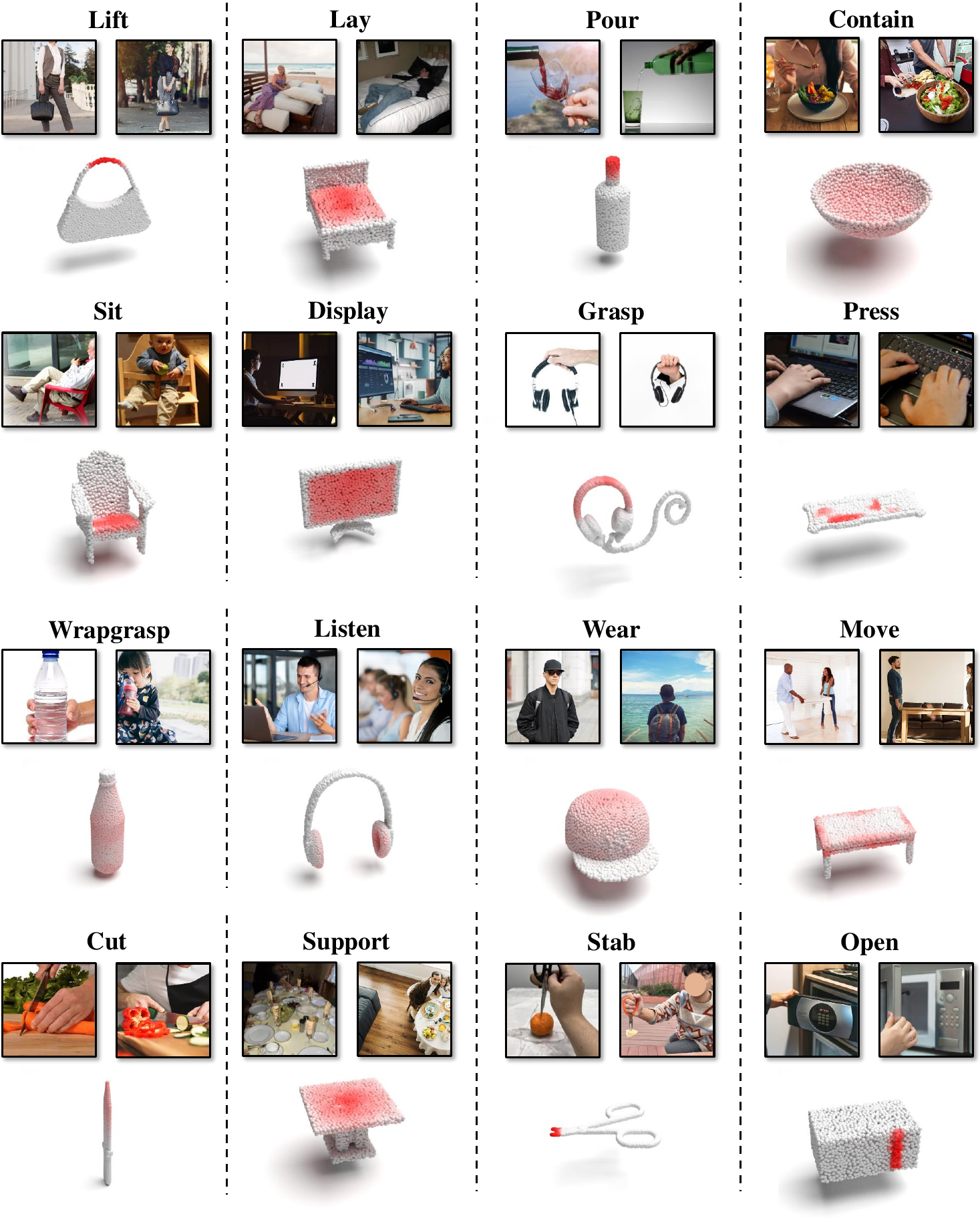}
    \caption{\textbf{Visualization of our method across different affordance categories in setting \texttt{seen}.}}
    \label{apdix:fig:ap_vis3}
\end{figure*}

\begin{figure*}[t]
    \centering
    \includegraphics[width=0.5\textwidth]{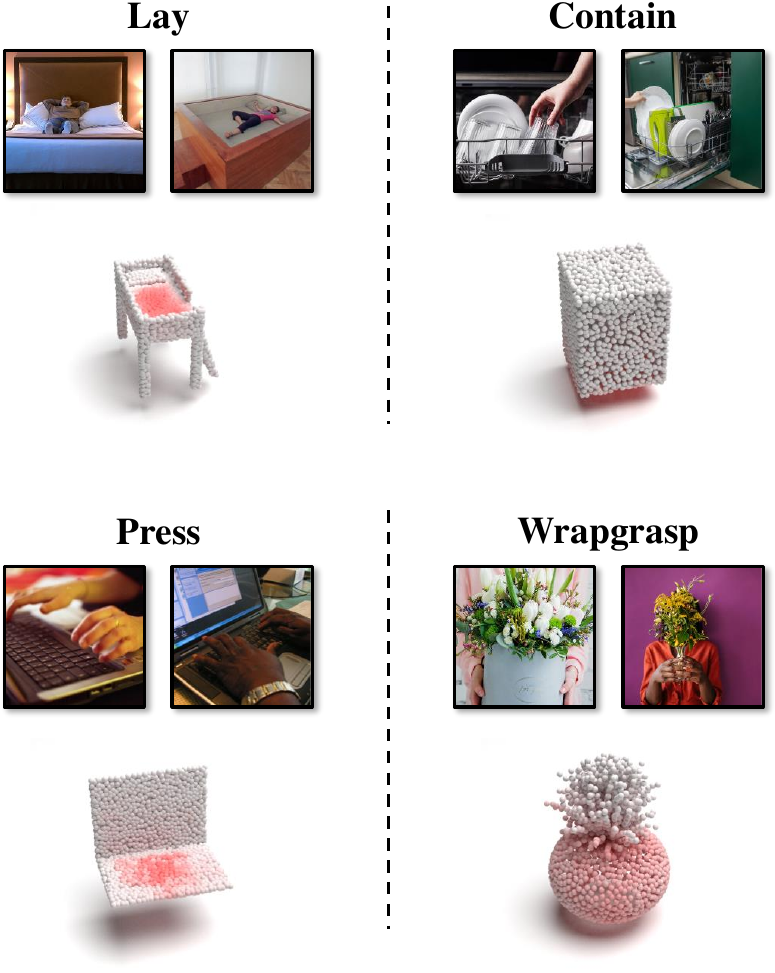}
    \caption{\textbf{Visualization of our method across different affordance categories in setting \texttt{unseen}.}}
    \label{apdix:fig:ap_vis4}
\end{figure*}

We show the evaluation results on our \dataset of every affordance category in both settings of \textbf{\texttt{seen}} (Table 5) and \textbf{\texttt{unseen}} (Table 6). Moreover, the additional qualitative results are shown in~\cref{apdix:fig:ap_vis1,apdix:fig:ap_vis2,apdix:fig:ap_vis3,apdix:fig:ap_vis4}.

\begin{table*}[ht]
\centering
\begin{tabular}{cr ccccccccc}
\toprule

\textbf{Affordances} & \textbf{Metrics} & \textbf{MBDF} & \textbf{PMF} & \textbf{FRCNN} & \textbf{ILN} & \textbf{PFusion} & \textbf{XMF} & \textbf{IAGNet} & \textbf{LASO} & \textbf{\ours} \\ \midrule
\multirow{4}{*}{\textbf{contain}}    & AUC $\uparrow$ & 84.64  & 73.86  & 70.30  & 75.93  & 71.34  & 77.84  & 78.07  & 66.07  & 81.06  \\
                                   & aIOU $\uparrow$ & 10.11   & 3.45   & 2.85   & 4.38   & 3.74   & 4.59  & 14.31  & 5.15  & 15.41  \\
                                   & SIM $\uparrow$ & 0.433  & 0.357  & 0.388  & 0.370  & 0.349  & 0.382  & 0.514  & 0.305  & 0.533  \\
                                   & MAE $\downarrow$ & 0.223  & 0.133  & 0.191  & 0.152  & 0.132  & 0.149  & 0.103  & 0.123  & 0.104  \\ \midrule

\multirow{4}{*}{\textbf{lay}}     & AUC $\uparrow$  & 85.38  & 85.14  & 82.41  & 83.51  & 84.75  & 83.82  & 92.46  & 83.30  & 88.93  \\
                                   & aIOU $\uparrow$ & 18.51   & 11.46   & 8.49   & 12.87   & 13.11   & 10.75   & 24.53   & 15.93   & 23.85  \\
                                   & SIM $\uparrow$  & 0.463  & 0.476  & 0.379  & 0.465  & 0.495  & 0.447  & 0.659  & 0.496  & 0.372  \\
                                   & MAE $\downarrow$ & 0.228  & 0.146  & 0.210  & 0.158  & 0.141  & 0.162  & 0.091  & 0.120  & 0.046  \\ \midrule
\multirow{4}{*}{\textbf{sit}}   & AUC $\uparrow$ & 70.88  & 73.73  & 72.26  & 73.34  & 74.45  & 73.97  & 96.27  & 68.30  & 72.18  \\
                                   & aIOU $\uparrow$ & 8.17   & 6.29   & 4.88   & 7.67   & 7.77   & 6.22  & 36.36  & 5.41  & 13.52  \\
                                   & SIM $\uparrow$ & 0.381  & 0.415  & 0.361  & 0.415  & 0.428  & 0.400  & 0.724  & 0.361  & 0.518  \\
                                   & MAE $\downarrow$  & 0.253  & 0.142  & 0.203  & 0.156  & 0.140  & 0.159  & 0.065  & 0.130  & 0.110  \\ \midrule
\multirow{4}{*}{\textbf{wrapgrasp}}   & AUC $\uparrow$ & 40.90  & 43.72  & 51.39  & 45.15  & 42.80  & 44.15  & 96.27  & 43.85  & 72.18  \\
                                   & aIOU $\uparrow$ & 2.28   & 1.56   & 1.58   & 2.02   & 1.35   & 1.63  & 36.36  & 1.42  & 13.52  \\
                                   & SIM $\uparrow$ & 0.465  & 0.463  & 0.550  & 0.466  & 0.443  & 0.493  & 0.724  & 0.364  & 0.518  \\
                                   & MAE $\downarrow$  & 0.223  & 0.151  & 0.166  & 0.159  & 0.155  & 0.155  & 0.065  & 0.171  & 0.110  \\ \midrule
\multirow{4}{*}{\textbf{open}}   & AUC $\uparrow$ & 55.78  & 69.13  & 64.86  & 72.56  & 67.38  & 62.86  & 96.27  & 67.17  & 72.18  \\
                                   & aIOU $\uparrow$ & 1.21   & 1.06   & 0.885   & 1.51   & 0.865   & 0.822  & 36.36  & 5.24  & 13.52  \\
                                   & SIM $\uparrow$ & 0.119  & 0.125  & 0.110  & 0.129  & 0.110  & 0.110  & 0.724  & 0.128  & 0.518  \\
                                   & MAE $\downarrow$  & 0.171  & 0.119  & 0.222  & 0.154  & 0.114  & 0.159  & 0.065  & 0.077  & 0.110  \\ \midrule

\multirow{4}{*}{\textbf{display}}   & AUC $\uparrow$ & 36.27  & 34.39  & 28.50  & 39.17  & 37.18  & 37.59  & 91.46  & 39.41  & 72.18  \\
                                   & aIOU $\uparrow$ & 4.48   & 2.40   & 2.99   & 2.69   & 2.12   & 2.86  & 29.01  & 2.03  & 13.52  \\
                                   & SIM $\uparrow$ & 0.139  & 0.146  & 0.160  & 0.150  & 0.138  & 0.161  & 0.674  & 0.096  & 0.518  \\
                                   & MAE $\downarrow$ & 0.343  & 0.233  & 0.281  & 0.239  & 0.212  & 0.243  & 0.083  & 0.186  & 0.110  \\ \midrule
\multirow{4}{*}{\textbf{stab}}     & AUC $\uparrow$ & 56.80  & 82.48  & 85.35  & 68.19  & 89.11  & 82.00  & 33.02  & 60.79  & 88.93  \\
                                   & aIOU $\uparrow$ & 1.21   & 4.98   & 0.94   & 4.75   & 5.78   & 2.317   & 3.07   & 0.16   & 23.85  \\
                                   & SIM $\uparrow$  & 0.119  & 0.164  & 0.150  & 0.144  & 0.170  & 0.153  & 0.096  & 0.088  & 0.372  \\
                                   & MAE $\downarrow$ & 0.171  & 0.146  & 0.245  & 0.099  & 0.098  & 0.186  & 0.133  & 0.064  & 0.046  \\ \midrule
\multirow{4}{*}{\textbf{grasp}}     & AUC $\uparrow$ & 51.08  & 56.05  & 53.01  & 58.32  & 50.61  & 50.07  & 33.02  & 47.45  & 88.93  \\
                                   & aIOU $\uparrow$ & 0.86   & 0.92   & 0.99   & 0.82   & 0.717   & 0.80   & 3.07   & 0.61   & 23.85  \\
                                   & SIM $\uparrow$  & 0.467  & 0.430  & 0.457  & 0.370  & 0.384  & 0.439  & 0.096  & 0.178  & 0.372  \\
                                   & MAE $\downarrow$ & 0.155  & 0.144  & 0.198  & 0.123  & 0.125  & 0.160  & 0.133  & 0.147  & 0.046  \\ \midrule
\multirow{4}{*}{\textbf{press}}     & AUC $\uparrow$ & 88.78  & 89.19  & 90.46  & 89.09  & 89.64  & 88.85  & 33.02  & 87.80  & 88.93  \\
                                   & aIOU $\uparrow$ & 10.93   & 4.60   & 3.60   & 6.06   & 5.91   & 4.20   & 3.07   & 11.24   & 23.85  \\
                                   & SIM $\uparrow$  & 0.263  & 0.197  & 0.160  & 0.211  & 0.231  & 0.181  & 0.096  & 0.403  & 0.372  \\
                                   & MAE $\downarrow$ & 0.250  & 0.166  & 0.237  & 0.172  & 0.137  & 0.189  & 0.133  & 0.088  & 0.046  \\ \midrule
\multirow{4}{*}{\textbf{cut}}     & AUC $\uparrow$ & 51.39  & 78.70  & 79.06  & 59.00  & 67.53  & 69.82  & 33.02  & 63.10  & 88.93  \\
                                   & aIOU $\uparrow$ & 1.19   & 4.62   & 1.95   & 2.71   & 3.75   & 3.07   & 3.07   & 2.31   & 23.85  \\
                                   & SIM $\uparrow$  & 0.167  & 0.253  & 0.219  & 0.192  & 0.205  & 0.214  & 0.096  & 0.213  & 0.372  \\
                                   & MAE $\downarrow$ & 0.183   & 0.150  & 0.234  & 0.123  & 0.125 & 0.184  & 0.133  & 0.090  & 0.046  \\ \bottomrule

\end{tabular}
\caption{\textbf{Affordance prediction metrics of every affordance category on \dataset in the \texttt{unseen} setting.}}
\label{apdix:tab:res_2}
\end{table*}

\section{Additional Discussion on Limitations}
\label{apdix:sec:limitation}

In this section, we delve deeper into the limitations of our current approach, particularly focusing on the aspects that may hinder its applicability to real-world scenarios and embodied manipulation tasks.

\begin{itemize}
    \item \textbf{Simplicity of 3D Objects in the Dataset}
    
    While our dataset has been instrumental in advancing the understanding of affordance grounding, it is important to recognize that the 3D objects included are relatively simple compared to those encountered in real-world environments. The objects in our dataset generally lack the complexity of everyday items, such as intricate geometries, fine details, and diverse material properties. This simplicity could lead to an overestimation of our model's effectiveness when applied to more complex, real-world objects.

    \item \textbf{Focus on Visual Understanding}
    
    Our current framework is primarily designed for the visual understanding of affordances, focusing on predicting functional regions based on visual cues. Our approach does not take into account important factors related to manipulation, such as the size of the manipulator, the direction and force of the manipulation action, and the physical properties of the objects (\eg, weight, friction, or flexibility).
\end{itemize}

Our future work aims to incorporate more complex and varied 3D objects and more manipulation-specific features to better reflect the challenges posed by real-world applications.

\end{document}